\documentclass[sigconf]{acmart}
\usepackage{geometry}
 \usepackage{graphicx}
 \graphicspath{{../pdf/}{../jpeg/}}
   \DeclareGraphicsExtensions{.pdf,.jpeg,.png}

\usepackage{amsmath}
\usepackage{cases}

\usepackage[caption=false,font=small]{subfig}
\usepackage{layouts}
\usepackage[export]{adjustbox}

\usepackage{array}
\usepackage{diagbox}
\usepackage{multirow}
\usepackage{booktabs}
\usepackage[bottom]{footmisc}
\usepackage{varwidth}

\usepackage[labelfont=bf, textfont=small, skip = 2pt]{caption}
\usepackage[rightcaption]{sidecap}
\usepackage{setspace}
\usepackage{enumitem}

\usepackage{float}

\usepackage[normalem]{ulem}
\usepackage{notoccite}
\usepackage{nopageno}

\expandafter\def\expandafter\normalsize\expandafter{%
    \normalsize
    \setlength{\abovedisplayskip}{1pt plus 1pt minus 1pt}
    \setlength{\belowdisplayskip}{0.75pt plus 1pt minus 0pt}
    \setlength{\abovedisplayshortskip}{0pt plus 0pt minus 1pt} % short skip for equations that start after the ending of the previous line
    \setlength{\belowdisplayshortskip}{0.75pt plus 1pt minus 1pt}
}

\DeclareCaptionOption{tight}[]{%
\captionsetup{farskip=1pt,topadjust=0pt,captionskip=2pt,nearskip=0pt,margin=0pt}
}
\begin{document}

%\copyrightyear{2018} 
%\acmYear{2018} 
%\setcopyright{acmcopyright} % Copyright
%\acmConference[ISLPED '18]{ISLPED '18: International Symposium on Low Power Electronics and Design}{July 23--25, 2018}{Seattle, WA, USA}
%\acmBooktitle{ISLPED '18: ISLPED '18: International Symposium on Low Power Electronics and Design, July 23--25, 2018, Seattle, WA, USA}
%\acmPrice{15.00}
%\acmDOI{10.1145/3218603.3218616}
%\acmISBN{978-1-4503-5704-3/18/07}

%%% Following 4 commands for arXiv submission only
\setcopyright{none}
\acmConference{}{}{}
\settopmatter{printacmref=false} % Removes citation information below
\renewcommand\footnotetextcopyrightpermission[1]{} % removes footnote with conference information in first column

\author{Ankit Mondal}
%\authornote{Dr.~Trovato insisted his name be first.}
%\orcid{1234-5678-9012}
\affiliation{%
  \institution{University of Maryland}
  %\streetaddress{P.O. Box 1212}
  \city{College Park}
  \state{Maryland}
  \postcode{20742}
}
\email{amondal2@terpmail.umd.edu}

\author{Ankur Srivastava}
\affiliation{%
  \institution{University of Maryland}
  \city{College Park}
  \state{Maryland}
  \postcode{20742}
}
\email{ankurs@umd.edu}

\title{In-situ Stochastic Training of MTJ Crossbar based Neural Networks}

\begin{abstract}
    Owing to high device density, scalability and non-volatility, Magnetic Tunnel Junction-based crossbars have garnered significant interest for implementing the weights of an artificial neural network. The existence of only two stable states in MTJs implies a high overhead of obtaining optimal binary weights in software. We illustrate that the inherent parallelism in the crossbar structure makes it highly appropriate for in-situ training, wherein the network is taught directly on the hardware. It leads to significantly smaller training overhead as the training time is independent of the size of the network, while also circumventing the effects of alternate current paths in the crossbar and accounting for manufacturing variations in the device. We show how the stochastic switching characteristics of MTJs can be leveraged to perform probabilistic weight updates using the gradient descent algorithm. We describe how the update operations can be performed on crossbars both with and without access transistors and perform simulations on them to demonstrate the effectiveness of our techniques. The results reveal that stochastically trained MTJ-crossbar NNs achieve a classification accuracy nearly same as that of real-valued-weight networks trained in software and exhibit immunity to device variations.
\end{abstract}

%%%%


\maketitle

\section{Introduction}
Deep Neural Networks (DNNs) have become a popular choice for tasks such as image classification, face recognition, and Natural Language Processing. This has however been at the cost of massive computations on von Neumann architectures exhibiting high energy and area requirements \cite{dean2012large}. The emergence of novel devices and special-purpose architectures has called for a shift from conventional digital hardware for implementing neural algorithms \cite{suri2013bio}.

Attempts have been made towards dedicated hardware designs and realization of the synaptic weights (and neurons) of a Neural Network (NN) by using CMOS transistors in an analog fashion \cite{misra2010artificial}; but these have met with challenges of scalability and volatility. %\textit{(the latter requiring constant training)}. 
Parallel research work has focused on using post-CMOS devices such as memristors, which are non-volatile devices with a variable resistance \cite{prezioso2015training}. However, the fabrication of multilevel memristors with stable states is still a challenge \cite{zhang2016stochastic}.
%\textit{Their use as synapse leverages the property that measured voltage pulses can change their resistance.} 
%Further, the possibility of designing high-density and low power memristive crossbars have paved the way for efficient neural computations, including the implementation of learning algorithms.

Another choice is the Magnetic Tunnel Junction (MTJ), an emerging \textit{binary} device (since it has 2 stable states) which has shown its potential as storage elements and is a promising candidate for replacing CMOS in memory chips \cite{wang2013low}. Its non-volatility and scalability makes it a particularly lucrative choice for logic-in-memory type architectures for neural networks. MTJs and memristors can be connected in a crossbar configuration which allows greater scalability and higher performance due to their inherent parallelism \cite{prezioso2015training,zhang2016all,vincent2015spin}. Several studies have investigated how the crossbar arrays with memristors \cite{querlioz2013immunity,saighi2015plasticity}, MTJs \cite{zhang2016stochastic,vincent2015spin} and domain-wall ferromagnets \cite{sengupta2016hybrid, saighi2015plasticity} can implement Spiking Neural Networks (SNN) trained using Spike-Timing Dependent Plasticity (STDP), both . %\textit{This has been done with the aim to mimic the learning process in biological NNs.}
Hasan et al. \cite{hasan2014enabling} and Soudry et al.\cite{soudry2015memristor} have implemented multi-layer NNs on memristive crossbars trained on-chip using the backpropagation algorithm and demonstrated on supervised learning tasks. 
%\textit{The former emphasize the need for on-chip training to overcome the effects of device variability and alternate current paths in the crossbar.}
%Also, the parallelism offered by the crossbar configuration allows high performance of the model. 

Continuous weight networks can be simplified into discrete weight networks without significant degradation in classification accuracy while achieving substantial power benefits \cite{rastegari2016xnor}. The use of discrete weight networks, such as BinaryConnect \cite{courbariaux2015binaryconnect} and in \cite{li2016ternary}, also stems from the challenge to address the high storage and computational demands of a large number of full-precision weights.
%\textit{It helps get rid of the multiplications in both the inference and programming phases, replacing them with additions instead.}
The existence of only 2 stable states in MTJs makes them a good candidate for the realization of binary weight networks. One way of training such NNs is to perform weight updates stochastically, which is justifiable from evidences that learning in human brains also has some stochasticity associated \cite{suri2013bio}. That such a method can lead to convergence with high probability in a finite time has been shown in \cite{senn2005convergence}. 
%In \cite{suri2013bio}, Suri et al. propose the use of conductive-bridge RAMs as binary synapses and demonstrate probabilistic STDP learning rules on CBRAM architectures with and without selector devices. 

Obtaining optimal weights for a binary network in software can be impractical because its discrete nature requires integer programming. Also, when physically realizing an NN on hardware, the underlying device variations can have a substantial impact on the model accuracy, and need to be accounted for in the training process. Merely characterizing the variations in the hardware platform is not sufficient for overcoming this issue.
%A naive approach would be to first characterize the variations in the hardware platform, followed by the software-based approach to training the NN which captures the device variability characteristics. Needless to say, such an approach would not only be slow but also inaccurate.

In this paper, we explore the use of MTJ crossbars for the hardware implementation of the synaptic weight matrices of a neural network. We propose the \textit{in-situ} training of such an MTJ crossbar NN, which allows us to exploit its inherent parallelism for significantly faster training and also accounts for device variations.
%For example, we illustrate that when trained using backpropagation, time complexity for the weight updates is independent of the size of the network.
 We advocate a probabilistic way of updating the MTJ synaptic weights through the gradient descent algorithm by exploiting the stochasticity in their switching. We experiment with two crossbar structures: with and without access transistors. The latter poses the additional challenge of sneak-path currents during programming which makes training in-situ the only choice to achieve satisfactory performance. Finally, we support our proposed techniques with data by modeling device and circuit properties and running simulations.

%Obtaining optimal binary weights for an NN is an NP-hard problem, and this implies the need for stochastic training. One may obtain optimal binary weights and attempt to program synapses on a crossbar individually. Although such a process would be successful on a 1T1R crossbar, the presence of sneak paths in a 1R crossbar would render it ineffective.

%In this paper, we explore the possibility of using MTJ crossbars for the hardware implementation of the synaptic weight matrices of a neural network and argue that issues such as leakage currents and device variations make it necessary to train them in-situ. We propose a probabilistic way of updating the MTJ synaptic weights by exploiting the stochasticity in their switching, and map parameters of the gradient descent algorithm to those controlling their switching probability. Then we describe crossbar architectures with and without selector devices,  and demonstrate how they can be stochastically trained in-situ, keeping in mind effects of sneak paths in the latter. 

\vspace{-2mm}
\section{Background}

In this section we describe the basics of neural networks and the parallelism offered by the crossbar architecture, and introduce the characteristics of Magnetic Tunnel Junctions.

\vspace{-2mm}
\subsection{Neural Networks} \vspace{-0.5mm}
The computation performed by any layer of an NN during the inference (forward propagation) phase basically comprises a matrix-vector multiplication. Say, $x\in R^M$ is the input to a layer and $W\in R^{N\times M}$ represents the synaptic weight matrix, then the output $y\in R^N$ is
\begin{equation}
    y = f(Wx)
    \label{inf1}
\end{equation}

\noindent where f() is an activation function. 
%Effectively, each neuron output involves a dot product, with the $j^{th}$ one being
%\begin{equation}
%    y_j = f\left(\sum_{i=1}^{M} W_{ji}x_i \right)
 %   \label{inf2}
%\end{equation}
Training of the NN can be done by backpropagation using the \textit{gradient descent} optimization method. The weight update of the synapse connecting the $i^{th}$ input to the $j^{th}$ output is given as 
\begin{equation}
    \Delta W_{ji} = - \eta \frac{\partial E}{\partial W_{ji}} = - \eta x_i \delta_j
    \label{bp1}
\end{equation}

\noindent where E is the cost function of the presented input sample $x$, $\eta$ is the learning rate and $\delta_j$ is the error calculated at the $j^{th}$ output using $y$ and the desired output. It is worth noting that such a weight update is local in nature, in that it depends only on the information available at the synapse - the input to it and the error at its output. The weight update of the entire matrix can thus be written as
\begin{equation}
    \Delta W = - \eta \delta x^T
    \label{bp2}
\end{equation}

The major computational cost of this algorithm comes from the $O(M.N)$ complexity of eqns. (\ref{inf1}) and (\ref{bp2}) whose implementation on general-purpose hardware requires time and memory of the same order, thereby not motivating their use for large-scale applications. 
Fortunately, the nature of computation in eqn. (\ref{inf1}) and the locality of weight update enable the design of highly parallel hardware that reduce the overall complexity to O(1).

\vspace*{-2mm}
\subsection{The Crossbar Architecture} \vspace{-0.5mm}

\begin{figure}[b]
\centering
\captionsetup[subfloat]{font={stretch=0.5}}
\begin{minipage}{.35\textwidth}
\subfloat[Structure of an $M\times N$ crossbar]
{
\includegraphics[scale=0.37, trim = 8cm 3.5cm 0cm 3cm , clip, valign=t]{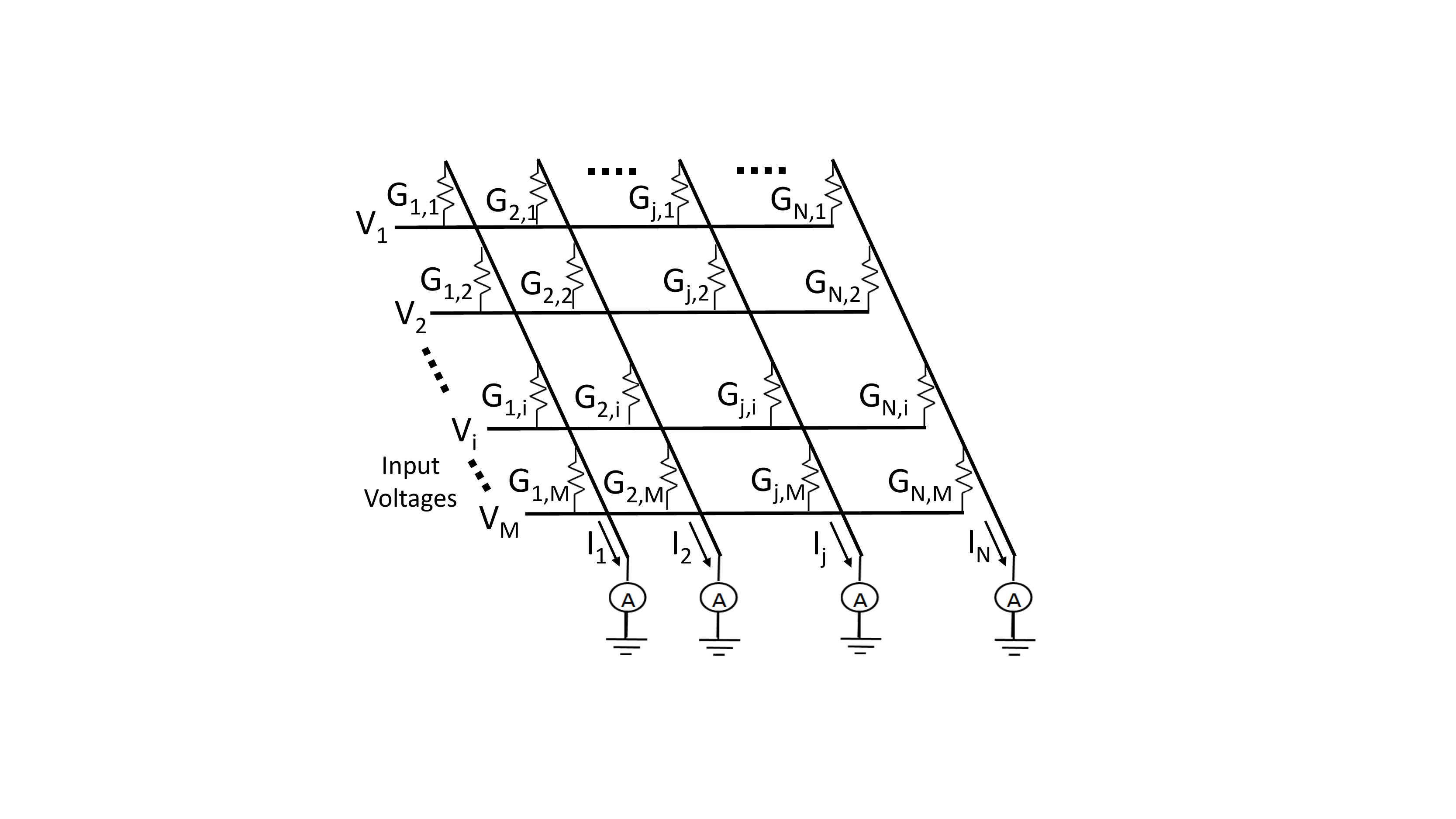}
    % Fully trimmed
    \label{crossbar_general}
}
\end{minipage}
\hfill
\begin{minipage}{.13\textwidth}
\hspace{-5mm}
\captionsetup[subfloat]{font={stretch=0.5},margin=-1pt,indention=15pt}
\subfloat[The 2 stable states of an MTJ]
 {
    \includegraphics[scale=0.26,trim = 11.7cm 9.8cm 12.8cm 4.3cm , clip, valign=t ]{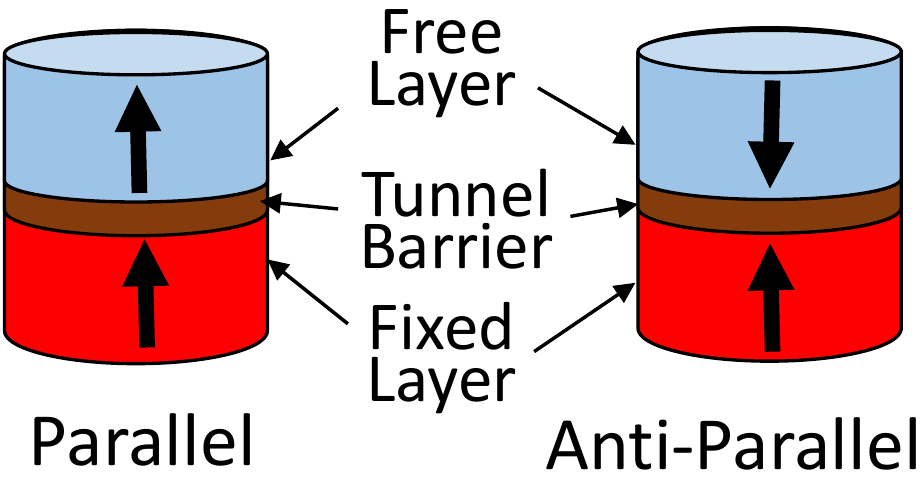}
    %Fully trimmed
    \label{MTJ_states}
}  \par
\subfloat[]
 {
\includegraphics[scale=0.25, trim = 10cm 3.3cm 12cm 5.5cm , clip]{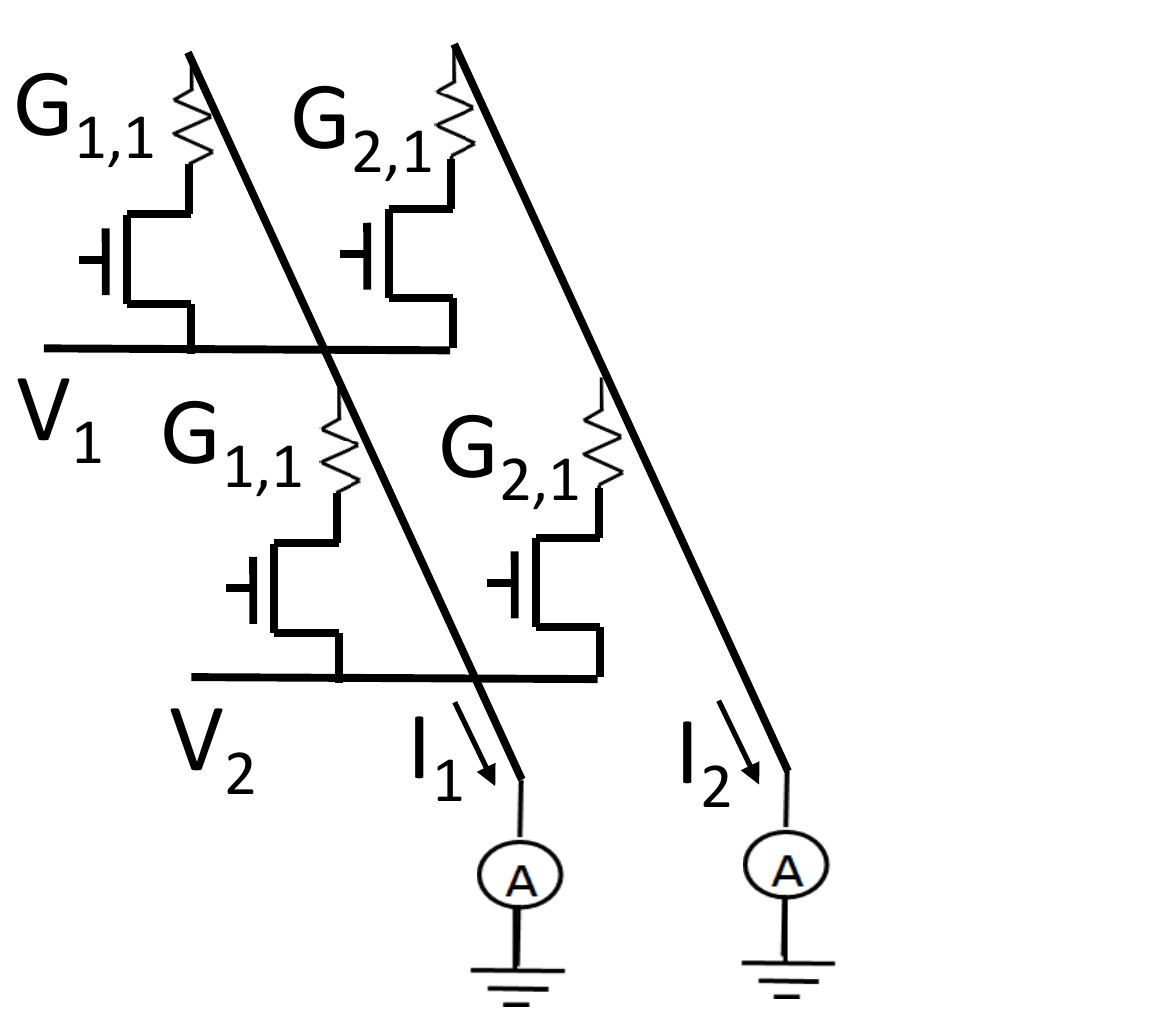}
% Fully trimmed
\label{crossbar_1t1r_general}
}
\end{minipage}
\caption{\small (a) A crossbar (b) Magnetic Tunnel Junction (c) A $2\times 2$ crossbar}
\label{crossbar_and_mtj}
\end{figure}

The physical realization of a synaptic weight matrix is possible using the grid-like crossbar structure where each junction has a resistance corresponding to one synapse. %\textit{This allows for the easy implementation of the forward and backward phases of the learning process as explained.}
Fig \ref{crossbar_and_mtj}\subref{crossbar_general} shows a simplified crossbar with each row corresponding to an input and each column to an output neuron. Let $V_i \in [-V,V]$ be the voltage applied at the $i^{th}$ input terminal and $G_{ji}$ be the conductance of the synapse connecting it to the $j^{th}$ output. By Ohm's Law, the current through that synapse is $G_{ji}V_i$ and by Kirchhoff's law the total current at the output is
\begin{equation}
    I_j  = \sum_{i} G_{ji}V_i
\end{equation}
\noindent which bears similarity to the dot products in (\ref{inf1}).
%and gives an output voltage $V_{o_{j}} = I_j R_j$
This can then be fed to suitable analog circuits for implementing the activation function.

Since the outputs are obtained almost instantaneously after the inputs are applied, the matrix-vector multiplication of eqn. (\ref{inf1}) is performed in parallel with constant time complexity. As for the update phase, the crossbar resistances can be modified by suitably modeling the required change as the product of 2 physical quantities derivable from the inputs and the errors. %\textit{In  the memristive crossbar of \cite{soudry2015memristor}, the update is done by using the property that the change in conductance is proportional to the product of the magnitude and duration of the applied voltage pulse.} 
In this way, the $O(M.N)$ operations can be done in parallel using the $M.N$ synapses.

\vspace*{-2mm}
\subsection{Magnetic Tunnel Junction} \vspace{-0.5mm}
The Magnetic Tunnel Junction (MTJ) 
%is an emerging spintronic device \textit{and a promising candidate for replacing CMOS as the basic element of future on-chip memory \cite{wang2013low}}. The MTJ
is a 2-terminal spintronic device consisting primarily of 2 ferromagnetic layers separated by a thin tunnel barrier (typically MgO). The magnetic orientation of one of the magnetic layers is fixed, whereas that of the other is free, as shown in fig \ref{crossbar_and_mtj}\subref{MTJ_states}. MTJs possess 2 stable states where the relative magnetic orientations of the \textit{free} and \textit{fixed} layers are Parallel (P) and Anti-Parallel (AP) respectively, with the P state exhibiting a lower resistance than the AP state ($R_P < R_{AP}$).
%It is this difference in resistance that allows a single-bit value to be encoded in the MTJ.
%and which is characterized by the Tunnel Magneto-Resistance, $TMR = (R_{AP} - R_{P})/R_{P}$.

It is possible to switch the state of the MTJ by passing spin-polarized current of appropriate polarity which flips the magnetization of the free layer through the mechanism of spin-transfer torque \cite{li2003magnetization}. The time required to switch is heavily dependent on the magnitude of the switching current. Not only that, this switching process is a stochastic one, in the sense that a pulse of given amplitude and duration has only a certain probability to successfully change the state. 
%\textit{This stochasticity does not arise from defects or variations in the device}, but
This stochasticity is due to thermal fluctuations in the initial magnetization angle and is an intrinsic property of the STT switching \cite{li2003magnetization}.

Depending on the magnitude I of the current and the critical current $I_{c0}$ \cite{zhang2016stochastic}, %given as .
the switching probability in the high-speed precessional regime $(I > I_{c0})$ is expressed as 
\begin{equation}
%\begin{aligned}
P(a,t) = exp(-4f(a)\Delta exp(-2t/T)), \ with \ 
f(a) = \left( \frac{2a}{a-1}\right)^{\left( \frac{-2}{a+1} \right)}
%and \ \ a = \frac{I}{I_{c0}}
%\end{aligned}
\label{sw_prob}
\end{equation}

%\begin{equation}
%I_{c0} = \frac{\alpha \gamma e \mu_0 M_s H_K V}{\mu_B \theta}
%\end{equation}

%\footnote{$H_{K}$ is the shape anisotropy field, $M_{s}$ is the saturation magnetization, $V$ is the volume of the free layer, $k_{B}$ is the Boltzmann constant, $\gamma$ is the gyromagnetic ratio, $\alpha$ is the damping constant, $\mu_0$ is the permeability of free space, $\theta$ is the spin transfer efficiency, $\mu_B$ is the Bohr magneton and $t_{p}$ is the pulse duration.}

%\noindent \textit{the switching characteristics can be divided into 3 categories: 1) Precessional switching, where $I > I_{c0}$ and switching typically takes place within 3ns, 2) Thermal Activation, with $I < 0.8I_{c0}$ and switching times $> 10 ns$, and 3) Dynamic Reversal, an intermediate regime very hard to model.} 

\noindent where $a=I/I_{c0}$, $t$ is the pulse width, $\Delta$ is the thermal stability and $T$ is the mean switching time (which is dependent on $a$)\cite{tomita2011high}. 

\begin{figure}
    \centering
    \subfloat[P vs t, for different values of I]
    {
        \includegraphics[scale=0.35,trim = 4.0cm 9.7cm 5.2cm 10.0cm , clip ]{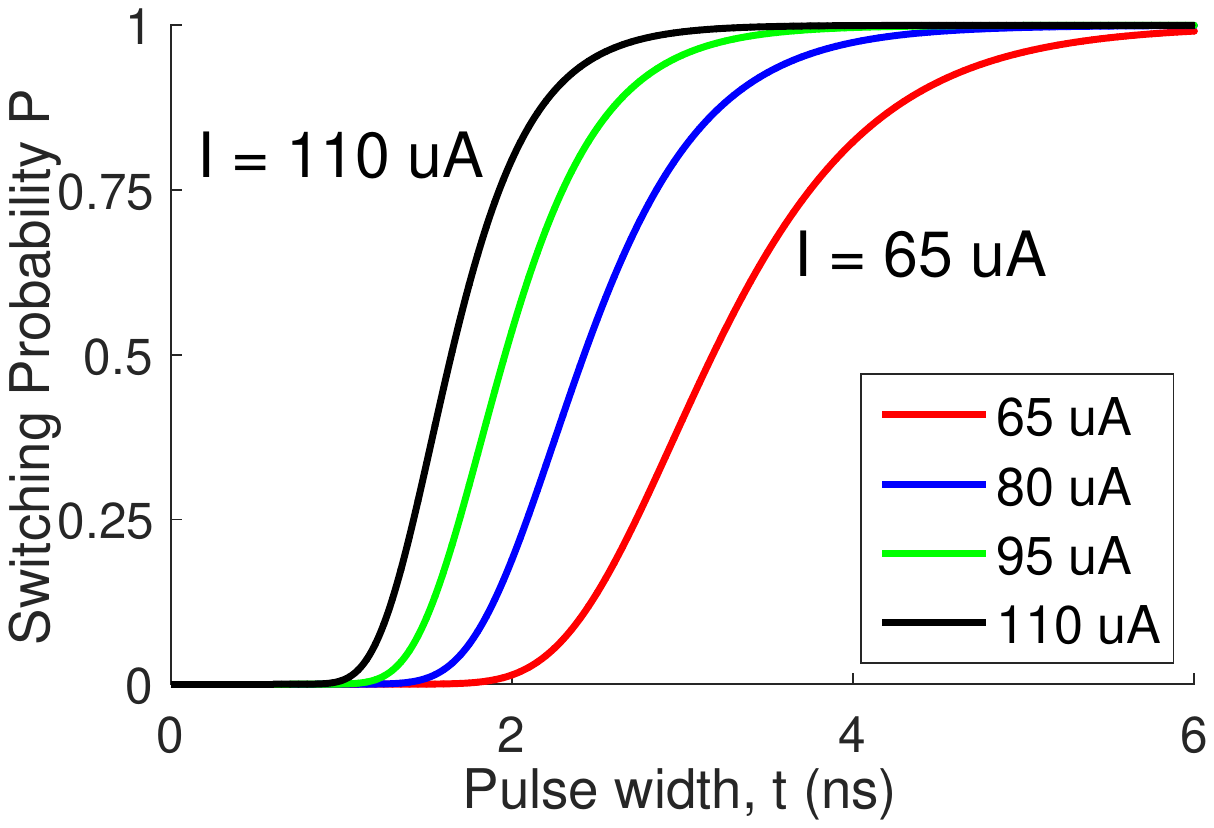}
        % Fully trimmed
        \label{sw_prob_t}
    }
    %\hfill
    \subfloat[P vs I, for different values of t]
    {
        \includegraphics[scale=0.39,trim = 5.4cm 9.8cm 6.3cm 10cm , clip ]{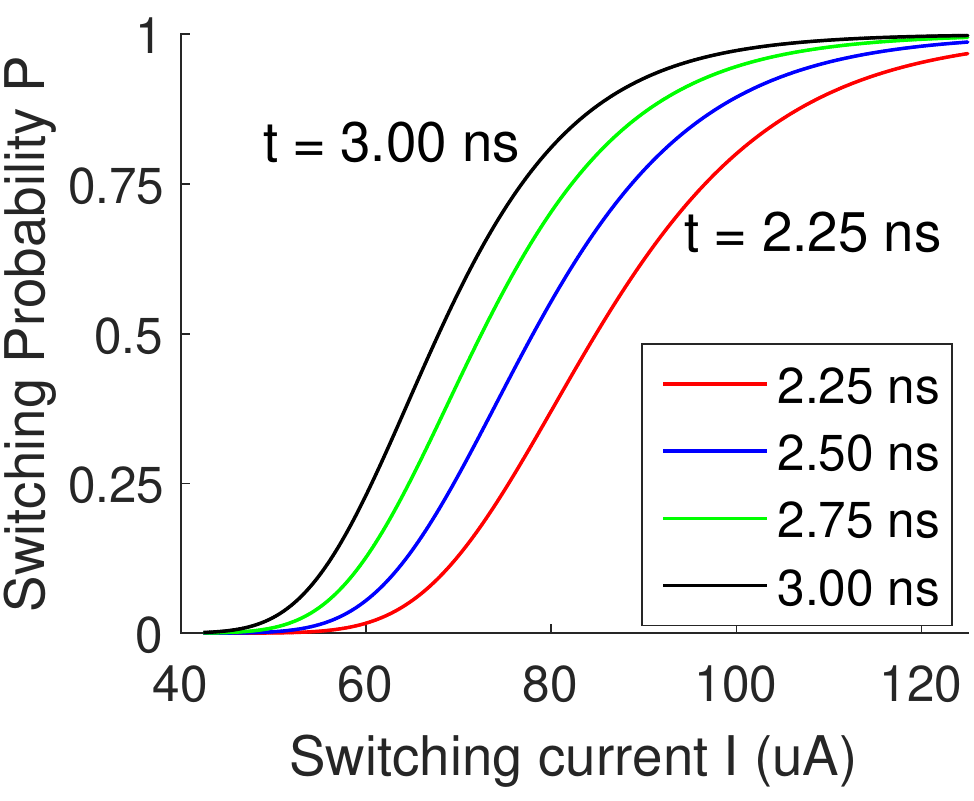}
        % Fully trimmed
        \label{sw_prob_I}
    }
    \caption{MTJ $AP\rightarrow P$ switching probability as a function of $t$ and $I$}
    \label{sw_prob_tandI}
\end{figure}

The spin transfer efficiency ($\theta$) of an MTJ is different for the 2 switching directions, with $\theta^{P\rightarrow AP}$ having a smaller value than $\theta^{AP\rightarrow P}$ \cite{zhang2012asymmetry}. This makes $I^{P\rightarrow AP}_{c0} > I^{AP\rightarrow P}_{c0}$, which means that the same magnitude and duration of current will correspond to different switching probabilities for the 2 switching directions. Fig. \ref{sw_prob_tandI} shows the dependence of the switching probability on pulse width and switching current for the $AP\rightarrow P$ transition. Observe the similarity in the nature of variation with $I$ and $t$. The $P\rightarrow AP$ transition too depicts this kind of a behavior, albeit with different values of $I$ and $t$.

\vspace{-2mm}
\section{MTJ Crossbar based Neural Networks}
The stochastic switching nature of MTJs has necessitated the usage of high write currents or write duration in memory applications to ensure low write errors. Alternatively, one can also use them to implement the synaptic weights in a crossbar where each cross-point would be an MTJ in one of its 2 states. 
%Although MTJs exhibit low $R_{OFF}/R_{ON}$ ratio, 
They are capable of being programmed with high speeds and exhibit endurance of the order of $10^{15}$ write cycles. 
However, the inherently binary nature of MTJs implies that such synapses can represent only 2 weight values and hence can implement only binary networks. Although it is possible to have some continuous behavior with the inclusion of a domain wall in the free layer \cite{sengupta2016hybrid}, the maturity of such technology is not at par with that of the binary version \cite{saighi2015plasticity}. 
%Substantial prior work has been done for synthesizing neural networks with binary weights, and has illustrated that their accuracy is very close to continuous weight networks at potentially much smaller implementation overhead \cite{courbariaux2015binaryconnect}.
%Such severe reduction in the precision of the weights will inevitably introduce high errors in network outputs and degrade its performance.

\textbf{Training Binary Networks:} Obtaining optimal binary weights for an NN is an NP-hard problem with an exponential time complexity, and hence a solution must involve training of the binary network of some form. This prompts the use of a probabilistic learning technique since the required weight update is continuous whereas any possible change in the conductance of the MTJ could only be discrete, in fact binary. As stated in \cite{suri2013bio}, stochastic update of binary weights is computationally equivalent to deterministic update of multi-level weights at the system level. 

In \cite{vincent2015spin}, Vincent et al. exploit the stochastic switching behavior of MTJs to propose its use as a "stochastic memristive synapse" in an SNN taught using a simplified STDP rule. However, there is no theoretical guarantee of the convergence of STDP for general inputs \cite{legenstein2005can}. We propose using a probabilistic learning approach by training using the gradient descent method (which requires weight updates of the form in eqn.  (\ref{bp1})) as demonstrated in section \ref{stochastic_update}
%Next, we demonstrate stochastic learning of an MTJ-based binary network in a supervised learning framework. 

\vspace{-2mm}
\subsection{The Motivation for In-situ Training}  \vspace{-0.5mm}
%As can be inferred from prior work, 
There are 2 ways (primarily) in which MTJs in the crossbar can be connected to their respective input and output terminals - 
\begin{enumerate}[leftmargin=*,topsep=1pt, partopsep=0pt, parsep=0pt, itemsep=0pt]
    \item With selector devices (1T1R) - Here each MTJ synapse is connected in series with an MOS transistor (as in fig. \ref{crossbar_and_mtj}\subref{crossbar_1t1r_general}), resulting in $O(M\times N)$ transistors in the crossbars. 
    %Literature in crossbar architectures have noted that the scaling of such a structure is limited by the presence of such access transistors.
    \item Without selector devices (1R) - Synapses are directly connected to the crossbar terminals; there are no transistors within the crossbar, such as the one in fig. \ref{crossbar_and_mtj}\subref{crossbar_general}. While a 1R structure provides greater scalability, it does so at the cost of reduced control of and access to individual synapses. 
\end{enumerate}
%\textit{It is the presence or absence of these transistors which dictates the level of access to the MTJ synapses during programming.} 
%We shall discuss the details of and the differences in these 2 architectures in the next section.
%One may obtain optimal binary weights and attempt to program synapses on a crossbar individually.
%One may question the need for on-chip training and argue that 

Stochastic learning can be done (simulated) offline and the final weights obtained can be programmed on to the crossbar deterministically. But, since MTJs have an inherently stochastic switching behaviour, deterministically programming them on a crossbar would require currents having high magnitude and duration to guarantee successful write operations. The possibility of selecting synapses to be written in the 1T1R architecture ensures no \textit{side-effects} of this method stemming from alternate current paths (because there would be none). But, despite circumventing this issue, this architecture can suffer from performance degradation due to the intrinsic device variations which only aggravate with scaling. On the other hand, in a 1R architecture, such high programming currents, when they sneak through alternate paths, are bound to cause unwanted changes in neighboring synapses owing to which the weights may never converge. This necessitates \textit{in-situ} training of the crossbar in probabilistic way for both 1T1R and 1R configurations, as only training on the hardware can account for both alternate paths and device variability.

\vspace{-2mm}
\subsection{Network Binarization} \vspace{-0.5mm}
\label{binarization}
%We can choose the binary weight values that the P and AP states of the MTJ represent. The quantization of real-valued weights to binary ones can introduce severe errors in the network output. Simply using $\pm$1 as the binary values is naive and estimating a good scaling factor $b$ is essential for the overall performance of the network. An appropriate way to determine a suitable $b$ is to minimize the L2 loss between the real-valued weights W and quantized ones,which amounts to solving the problem
%\begin{equation}
 %   b = \arg \min_{\alpha} \Vert W - \alpha \hat{W} \Vert \ \ subject \ to \ [\hat{W}]_{ij} \in \{1,-1\}
 %   \label{binarize}
%\end{equation}
Simply using $\pm$1 as the binary weight values, represented by the $P$ and $AP$ states of an MTJ, is naive and estimating a good scaling factor $b$ is essential for overall network performance. An appropriate way to determine a suitable $b$ is to minimize the L2 loss between the real-valued weights W and quantized ones, as was done in \cite{rastegari2016xnor}. 
%The discrete nature of the possible values that $\hat{W}$ can hold makes this an integer programming problem which may be impractical to solve for large $\hat{W}$. A simple and sensible abstraction to simplify this problem is to maintain the sign of the full-precision weights during binarization, i.e. choose $\hat{W} = sign(W)$. \textit{ This works well when the weights have mean 0 and a roughly symmetric distribution about it, which usually does happen to be the case.} 
This provides a solution $b = \Vert W \Vert_1 / n $ (the mean of absolute values of $W$). Thus an MTJ in the $P$ ($AP$) state would signify a weight of $+b$ ($-b$).

%\vspace{-2mm}
%\subsection{MTJ as a synapse} \vspace{-0.5mm}
%The weights of an NN are almost always bipolar whereas the conductance of an MTJ or memristor is always positive. \textit{ One way to implement a bipolar weight is to have it represented by a pair of conductances, $G^+$ and $G^-$, with the effective conductance being $G = G^+ - G^-$, and applying the same input voltage $V$ to both[]. The output current would then be obtained by subtracting the currents from these 2 branches, i.e. $I = I^+ - I^- = G^+V - G^-V = (G^+ - G^-)V$.} One method to realize negative weights is to offset the conductance with a fixed bias; in the case of MTJs, we choose $G_{bias} = (G_P + G_{AP})/2$ as it brings symmetry to the effective conductance: $G = G_P - G_{bias}$ would correspond to the positive weight, say $+b$, and $G = G_{AP} - G_{bias}$ would correspond to the negative weight, say $-b$. Thus, an MTJ in the P (AP) state would represent a weight of $+b (-b)$, and the output current in any column of the crossbar can be obtained by subtracting a bias current from the current received at the end of that column. That is, 
%\begin{equation}
%I = \sum_i (G_i - G_{bias})V_i = \sum_i (G_i V_i - G_{bias}V_i) = \sum_i (I_i - I_{bias,i}) 
%\end{equation}

\vspace{-2mm}
\section{In-situ Training of MTJ Crossbars}
\label{Crossbar_arch}
We first provide a high-level understanding of how an MTJ synaptic crossbar implementing an NN should work. For the sake of simplicity, all operations are described for a single-layer NN and can be easily scaled to multiple layers (more details subsequently). We then illustrate how the gradient descent method can be used for the stochastic weight update of MTJs, and finally describe the in-situ training procedure for the 2 crossbar architectures.

\vspace{-2mm}
\subsection{Overview of Operations} \vspace{-0.5mm}
 The training process is carried out as follows.
 
\textit{Read Phase}: Upon receiving a training input $x\in R^M$, the input terminals are applied with voltages $V^r_i \in \left[-V,V\right] \ \forall \ i$ proportional to $x_i$, whereas the output terminals are maintained at ground potential. Current $I_{ji} = G_{ji}V^r_i$ flows through the $(j,i)$ synapse and the total current $I$ at the output terminals are suitably converted to output $y$.

\textit{Write Phase}: Using $y$ and the desired output, calculate the error $\delta$. Table \ref{update_cases} lists the 4 possible cases of weight update depending on $x$ and $\delta$. The gradient descent algorithm requires a weight update of the form of eqn. (\ref{bp1}). An appropriate way to realize this, as suggested in \cite{lee2007defect}, is to set switching probabilities proportional to (the magnitude of) $\Delta w$ calculated in (\ref{bp1}). Our way of achieving this is explained next.

The process of read and write are carried out for each input sample and repeated for several iterations until convergence is achieved.

\begin{table}[t]
\captionsetup[table]{font={stretch=0.6}}
\setlength{\tabcolsep}{1.5mm}% Horizontal space between cell walls and cell text
\renewcommand{\arraystretch}{0.9} % Vertical width of a cell
\begin{tabular}{|c|c|c|c|c|}
    \hline
    Input & Error & $\Delta W$ & $W$ and $G$ & Switch \\
    \hline    
   $x>0$ & $\delta > 0$ & $\Delta W < 0$ & $Decreases$ & $P \rightarrow AP$ \\
   \hline
   $x>0$ & $\delta < 0$ & $\Delta W > 0$ & $Increases$ & $AP \rightarrow P$ \\
   \hline
   $x<0$ & $\delta > 0$ & $\Delta W > 0$ & $Increases$ & $AP \rightarrow P$ \\
   \hline
   $x<0$ & $\delta < 0$ & $\Delta W < 0$ & $Decreases$ & $P \rightarrow AP$ \\
   \hline
\end{tabular}
\captionof{table}{\textbf{Write phase.} Signs of $x$, $\delta$, and $\Delta W$, required change in weight W and conductance G, and the desired direction of switching of MTJ Synapse}
\label{update_cases}
\end{table}

\vspace{-2mm}
\subsection{Stochastic Learning of an MTJ Synapse} \vspace{-0.5mm}
\label{stochastic_update}
We will now describe how the stochasticity of MTJ switching can be used to perform weight updates with gradient descent method. Just as the weight update in eqn (\ref{bp1}) is a function of 2 variables (the input and the error), the probabilistic switching of MTJs can be controlled by 2 physical quantities- the magnitude and the duration of the programming current. We choose the magnitude of the write current to be dependent on the input $x_i$ and the duration on the error $\delta_j$. However, as can be evidenced from eqn (\ref{sw_prob}) and fig \ref{sw_prob_tandI}, the switching probability $P$ is a highly non-linear function of the parameters $a$ and $t$ (recall $a=I/I_{c0}$), whereas the desired probability, being proportional to $\Delta W_{ji}$, is a linear function of $x_i$ and $\delta_j$. Further, the switching probability does not immediately rise with the pulse width and the write current as they increase from 0, indicating some kind of soft threshold. Note that the direction of switching can be decided by the polarity of the write current.

We therefore model switching probabilities by a linear mapping of $x$ and $\delta$ to write current $I_{wr}$ and duration $t_{wr}$ respectively as follows. Usually $|x| \leq 1$, and henceforth assume for simplicity that $|\delta| \leq 1$ (can be ensured by normalizing and adjusting with $\eta$). The pulse width $t_{wr}$ is set at a minimum of $t_0$ and increases linearly with $|\delta|$ (since $t_{wr}$ needs to increase irrespective of the sign of $\delta$) as
\begin{equation}
t_{wr} = t_0 + t_1|\delta|
\label{write_time}
\end{equation}

\noindent Similarly, the write current ($I_{wr}$) would be a minimum of $I_0$ and increase linearly with $|x|$ as
\begin{equation}
I_{wr} = I_0 + I_1|x|
\label{write_current}
\end{equation}

\begin{figure}[b]
\begin{minipage}[b]{0.23\textwidth}
    %\centering
    \captionsetup[figure]{font={stretch=0.7}, parindent=10pt, margin={-20pt,0pt},indention=15pt}
    \includegraphics[scale = 0.42,trim = 11.5cm 4cm 12cm 6cm , clip]  {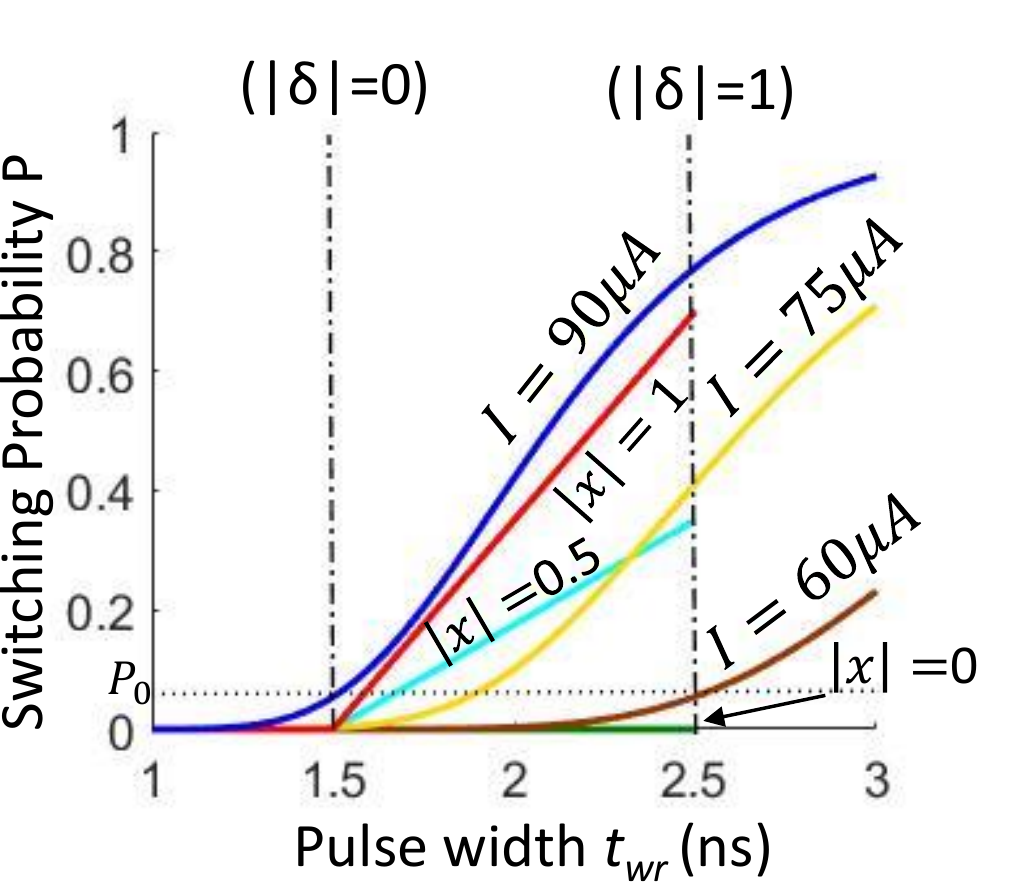}
    % Fully trimmed
    \captionof{figure}{P vs $t_{wr}$ of the linear model and desired probabilities (obtained with $\eta=0.7$) for $AP\rightarrow P$ transition. The region between the dashed vertical lines is of interest. The dark green, cyan and red straight lines plot desired probabilities for $|x|=0,0.5,$ and $1$ respectively. The brown, yellow and blue plots correspond to the actual switching probabilities (obtained from the linear model) for the mapped currents $I=60\mu A, 75\mu A,$ and $90\mu A$ }
    \label{Fit}
  \end{minipage}
  %\hfill
  \begin{minipage}[b]{0.23\textwidth}
    \centering
    \setlength{\tabcolsep}{1.5mm}% Horizontal space between cell walls and cell text
    \renewcommand{\arraystretch}{0.95} % Vertical width of a cell
    \captionsetup{skip=5pt, margin=6pt,font={stretch=0.9}}
    \begin{tabular}{|c|c|}
      \hline
      Weight & MTJ \\
      Update & Switching \\
      \hline
      $|\delta| = 0$ & $t_{wr} = t_0$ \\
      \hline
      $|\delta| = 1$ & $t_{wr} = t_0 + t_1$ \\
      \hline
      $|x| = 0$ & $I_{wr} = I_0$ \\
      \hline
      $|x| = 1$ & $I_{wr} = I_0 + I_1$ \\
      \hline
    \end{tabular}
    \captionof{table}{Boundary values of the parameters in the weight update eqn. (\ref{bp1}) and their counterpart in probabilistic switching of MTJ.}
    \label{boundary}
    
      \small
      \setlength{\tabcolsep}{0.5mm}% Horizontal space between cell walls and cell text
      \renewcommand{\arraystretch}{0.95} % Vertical width of a cell
      \captionsetup{skip=0pt}
      \begin{tabular}{|c|c|c|}
      \hline
      Direction & $AP\rightarrow P$ & $P\rightarrow AP$ \\
      \hline
      $t_0$ & $1.5 ns$ & $1.5 ns$ \\
      \hline
      $t_1$ & $1 ns$ & $1 ns$ \\
      \hline
      $I_0$ & $60 \mu A$ & $140 \mu A$ \\
      \hline
      $I_1$ & $30 \mu A$ & $60 \mu A$ \\
      \hline
    \end{tabular}
      \captionof{table}{The coefficients that fit the model for both $AP\rightarrow P$ and $P\rightarrow AP$ switching}
      \label{params}
    \end{minipage}
\end{figure}

We now wish to find coefficients $t_0, t_1, I_0$ and $I_1$ that yield MTJ switching probabilities ($P$) close to the desired probabilities of weight update. A certain probability of switching can be obtained for different combinations of $I$ and $t$, as is evident from fig. \ref{sw_prob_tandI}. We first fix the range of pulse widths by choosing suitable $t_0$ and $t_1$ (refer to table \ref{params}). We want a nearly 0 switching probability for $t_{wr} = t_0$ irrespective of the value of $I_{wr}$ because $\Delta W = 0$ for $\delta=0$ regardless of $x$. We thus choose the maximum $I_{wr}$ (which is $I_0 + I_1$) to be that value of $I$ for which the plot of P against $t_{wr}$ starts rising at $t_0$. That is 
\begin{numcases}{P(I_0 + I_1,t_{wr}) \ \ is }
    \nonumber
     < P_0 & for $t_{wr} < t_0$, \\
     \geq P_0 & for $t_{wr} \geq t_0$
\end{numcases}
where $P_0$ is a small value. So now even if $|x|$ is (as high as) $1$, $P= P_0$. In our experiments, we chose $P_0$ to be about $0.05$.

A symmetric argument holds when $x=0$. For $t_{wr}=t_0+t_1$, we want $P\approx 0$ if $I_{wr}=I_0$, (because $\Delta W = 0$ for $x=0$). But $P$ should start increasing as soon as $I_{wr}$ increases, that is

\begin{numcases}{P(I_{wr},t_0+t_1) \ \ is \ }
    \nonumber
    < P_0 & for $I_{wr} < I_0$ \\
    \geq P_0 & for $I_{wr} \geq I_0$
\end{numcases}

Fig \ref{Fit} shows how well the linear model approximates the required $AP\rightarrow P$ switching probabilities (similar curve fitting for $P\rightarrow AP$ as well). Table \ref{boundary} shows the write currents and duration for boundary values of $|x|$ and $|\delta|$ and table \ref{params} lists the values of the coefficients in eqns. (\ref{write_time}) and (\ref{write_current}). One could use non-linear models for mapping $|\delta|$ and $|x|$ to $t_{wr}$ and $I_{wr}$, respectively, in order to better fit the desired switching probabilities; however, that would complicate the analog circuit responsible for the conversion. Owing to this, and the closeness with which the linear model can replicate the stochastic switching characteristics, we stick to the linear version.

Next, we describe the 1T1R and 1R crossbar architectures implementing the NN. We show how these can be trained in-situ using the stochastic learning technique described above.

\vspace{-2mm}
\subsection{The 1T1R Architecture} 
\label{1t1r_arch}
\begin{figure}[b]
    %\centering
    \subfloat[An $M\times N$ crossbar]
    {
    \includegraphics[scale=0.23,trim = 2.7cm 11.7cm 7.2cm 2.8cm, clip]{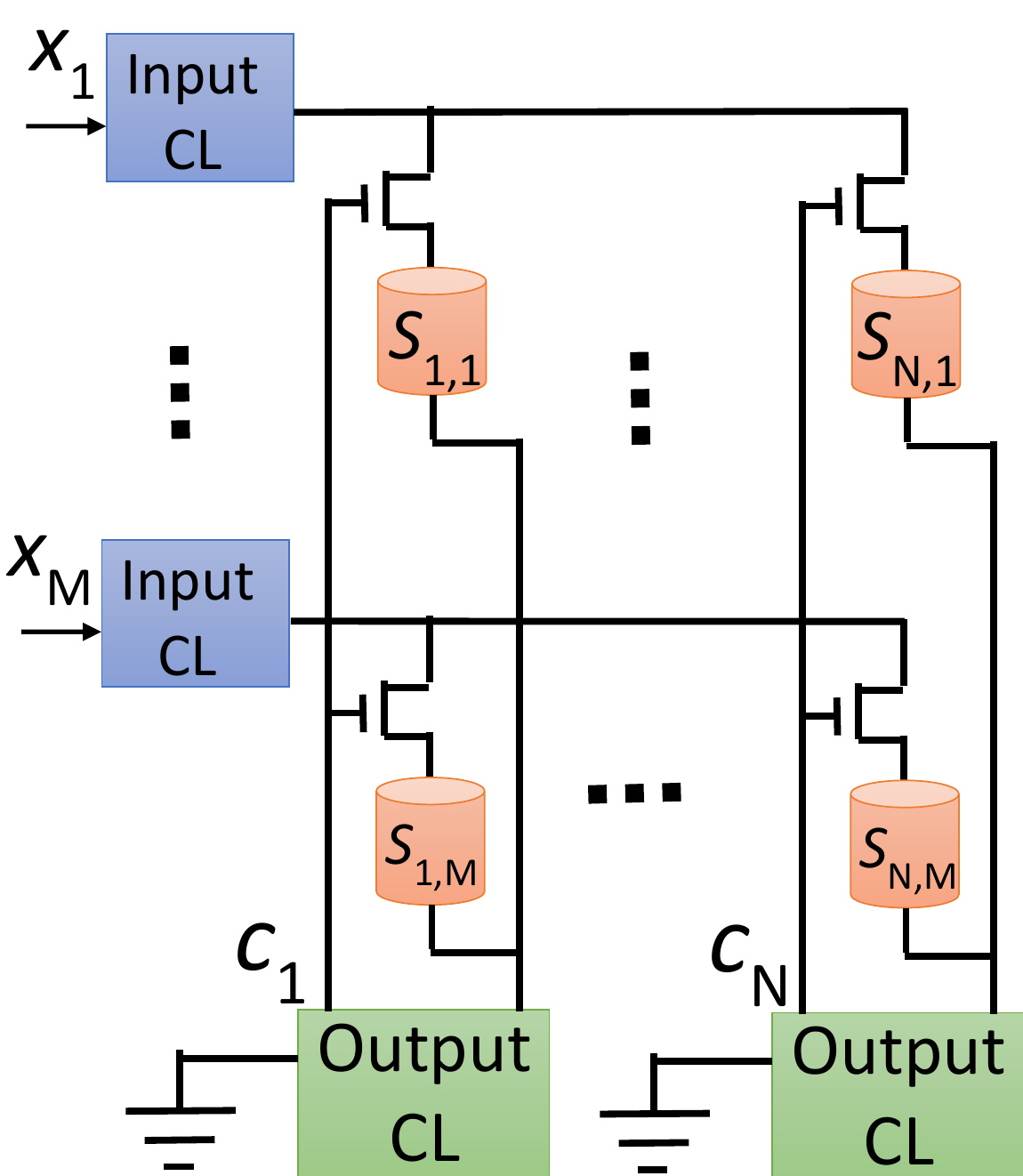}
    % Fully trimmed
    \label{crossbar_1t1r}
    }
    \quad
    \subfloat[Write voltages and control signals.]
    {
    \includegraphics[scale=0.25,trim = 0.2cm 12.8cm 0cm 2.8cm , clip]{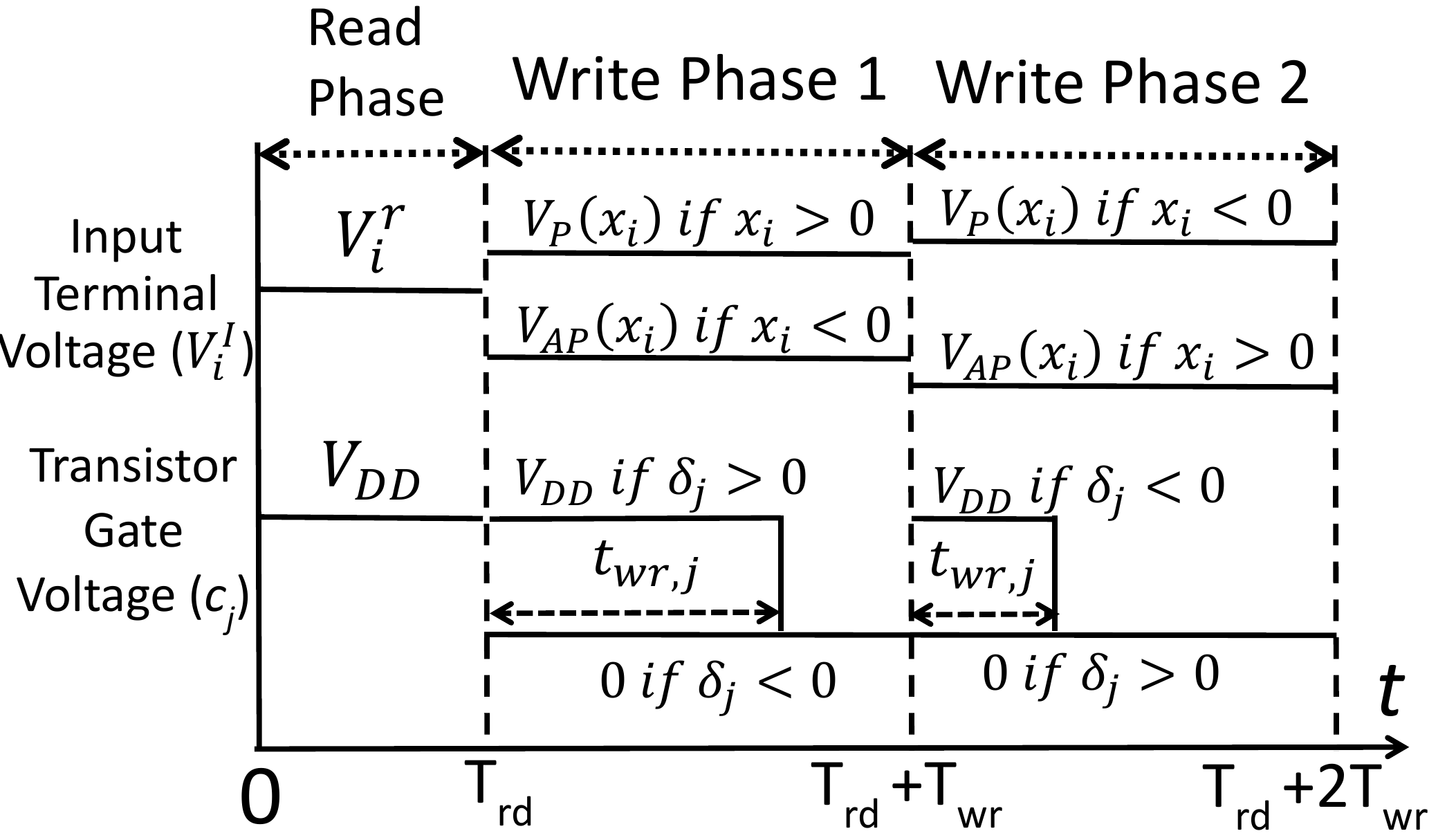}
    % Fully trimmed
    \label{signals_1t1r}
    }
    \caption{The 1T1R crossbar. (a) Schematic (b) Read \& write phases signals}
    \label{1t1r_figures}
\end{figure}

\vspace{-0.5mm}
This is the conventional architecture for memory applications where each cell has a selection transistor. One major advantage of being able to selectively turn off certain cells is that it disallows the presence of undesired sneak currents which lead to unnecessary power consumption at a minimum. Fig \ref{1t1r_figures}\subref{crossbar_1t1r} shows a 1T1R crossbar where each MTJ synapse is connected in series with an NMOS transistor. Input and output terminals are interfaced with necessary Control Logic (CL). All the transistors in a single column will have a common gate voltage since the corresponding synapses are connected to the same neuron output, and hence will always have the same error  \textquoteleft$\delta$\textquoteright \ and write pulse width $t_{wr}$.

Fig \ref{1t1r_figures}\subref{signals_1t1r} plots the signals during both the read and write phases. During the read phase ($0 \leq t \leq  T_{rd}$), all transistors are turned on: $c_j = V_{DD} \ \ \forall \ \ j=1...N$ so that all columns (neuron outputs) are read simultaneously. Inputs $x_i$ are provided to their respective input CLs which convert them to read voltages $V^r_i$. Output currents $I_j$ are processed by the output CLs. 

\vspace{-1mm} 
\subsubsection*{\textbf{Updating the crossbar:}}
Decide the write currents that should be provided to each input row and the pulse widths for each output column as described in sec. \ref{stochastic_update}. Recall that the former depend on $x$ and the latter on $\delta$. The direction of the currents would depend on the sign of the desired weight update. Apply suitable write voltages at the input terminals while grounding the output terminals to 0. 

For the $(j,i)$ synapse, the write pulse width depends on only $|\delta_j|$, and the write current magnitude depends on $|x_i|$. But the direction of switching depends on the signs of $\delta_j$ and $x_i$ (see Table \ref{update_cases}) and has to be decided by the polarity of current. For eg. two MTJ synapses belonging to the same row but different columns may have opposite signs of $\delta$. Thus, despite having the same input $x_i$, they are required to switch in opposite directions and hence need write voltages of opposite sign. This requires us to split the write phase into two parts as explained next.

Since the transistor gate control signals are connected to the output CLs, we can select or deselect a certain column based on information at its respective CL, which is the error $\delta$. We therefore program the crossbar sequentially in 2 stages, with the columns updated in a given stage depending on the signs of $\delta$. Each phase has a duration of $T_{wr}$ (which need not be more than $t_0 + t_1$, see eqn. (\ref{write_time})). The voltage signals in each phase are plotted in fig. \ref{1t1r_figures}\subref{signals_1t1r} and detailed below -
\begin{enumerate}[leftmargin=*,topsep=0pt, partopsep=0pt, parsep=0pt, itemsep=0pt]
    \item Phase 1: $T_{rd} \leq t \leq T_{rd} + T_{wr}$. Update the weights of the columns which had $\delta > 0$. Then, the transistor control signals would be
    \begin{numcases}{c_j =}
    \nonumber
    V_{DD}, & for $\delta_j > 0$ \textit{and} $0 \leq t - T_{rd}\leq t_{wr,j}$\\
    0, & for $\delta_j < 0$ \textit{or} $t_{wr,j} \leq t - T_{rd} \leq T_{wr}$
    \end{numcases}
    And the write voltages applied at the input terminals would be
    \begin{equation}
        V_{wr,i} = V_{P}(x_i)u(x_i) + V_{AP}(x_i)u(-x_i)
    \end{equation}
    where $u$ is the unit step function.
    %\begin{numcases}{V_{wr,i} =}
    %V_{P}(x_i), & for $x_i > 0$ \\
    %V_{AP}(x_i), & for $x_i < 0$
    %\end{numcases}
    
    \item Phase 2: $T_{rd} + T_{wr} \leq t \leq T_{rd} + 2T_{wr}$. Update the weights of those columns which had $\delta < 0$. Here, the signals are opposite to those in phase 1 as shown in fig. \ref{1t1r_figures}\subref{signals_1t1r}.
    
    %\begin{numcases}{V_{wr,i} =}
    %V_{AP}(x_i), & for $x_i > 0$ \\
    %V_{P}(x_i), & for $x_i < 0$
    %\end{numcases}
\end{enumerate}
Here $V_P$ $(V_{AP})$ is the voltage applied to switch from P$\rightarrow$AP (AP$\rightarrow$P) and can be obtained using (\ref{write_current}) and $R_P (R_{AP})$. $V_P$ and $V_{AP}$ still depend on $|x_i|$, but for brevity explicit mention will be omitted henceforth. Let MTJs in the crossbar be arranged in a way that positive (negative) current from the $i^{th}$ input terminal to $j^{th}$ output terminal can switch $S_{j,i}$ from $P\rightarrow AP$ ($AP\rightarrow P$); hence $V_P > 0$, $(V_{AP} < 0)$. Parameters in table \ref{params} give $V_P \in [0.68,0.98]$ volts and $V_{AP} \in [-0.81,-0.62]$ volts.

Thus we can see that the read and update operations are completed in $T_{rd} + 2T_{wr}$ time which is $O(1)$. %Despite its advantages, the scalability of the 1T1R configuration is limited by that of the transistor. 
Due to limitations on the scalability of 1T1R architecture, it is worth exploring the feasibility of transistor-less crossbars to achieve even higher density of integration.

\vspace{-2mm}
\subsection{The 1R Architecture} \vspace{-0.5mm}
\label{1R_arch}
Eliminating the need to have an access transistor for every synapse in the crossbar will allow for compact designs having an integration density of about $4F^2/$device. But the inability to select the synapses to be updated during programming results in leakage currents through alternate paths that not only waste energy but also can lead to undesirable changes in synaptic conductance. We first see the effect of such currents with the previously proposed write-strategy and then suggest a modified strategy (and circuit) for the 1R architecture

\vspace{-2mm}
\subsubsection{\textbf{Two-phase update:}} 
\label{two_phase_update}
\begin{figure}[t]
\centering
\captionsetup{font={stretch=0.5}}
\begin{minipage}{.14\textwidth}
\subfloat[]
 {
\includegraphics[height = 2.5cm, width = 1.00\textwidth, trim = 11.5cm 6.5cm 16cm 6cm, clip]{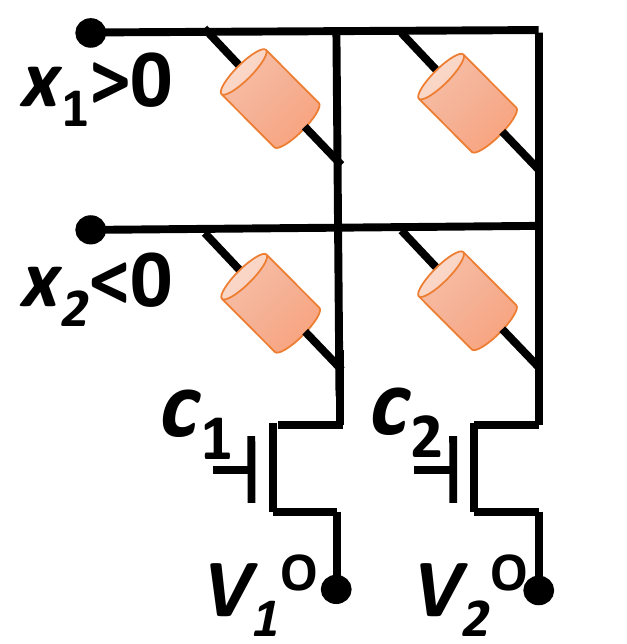}
\label{sneak_2phase_demo}
}  \par
\subfloat[]
 {
\includegraphics[height = 2.5cm, width = 1.00\textwidth, trim = 12cm 5cm 16cm 8cm , clip]{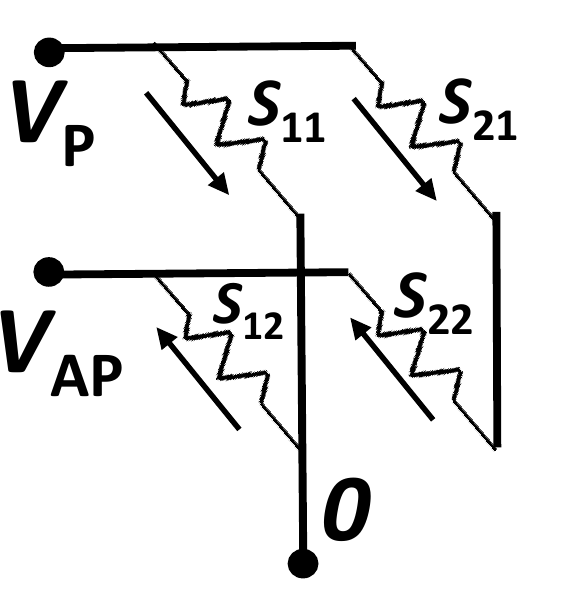}
\label{sneak_2phase_equiv}
}
\end{minipage}
\hfill
\begin{minipage}{.33\textwidth}
\centering
\subfloat[An $M\times N$ crossbar with 1R structure]
    {
    \includegraphics[scale=0.27,trim = 2cm 3.5cm 13.5cm 2.0cm , clip]{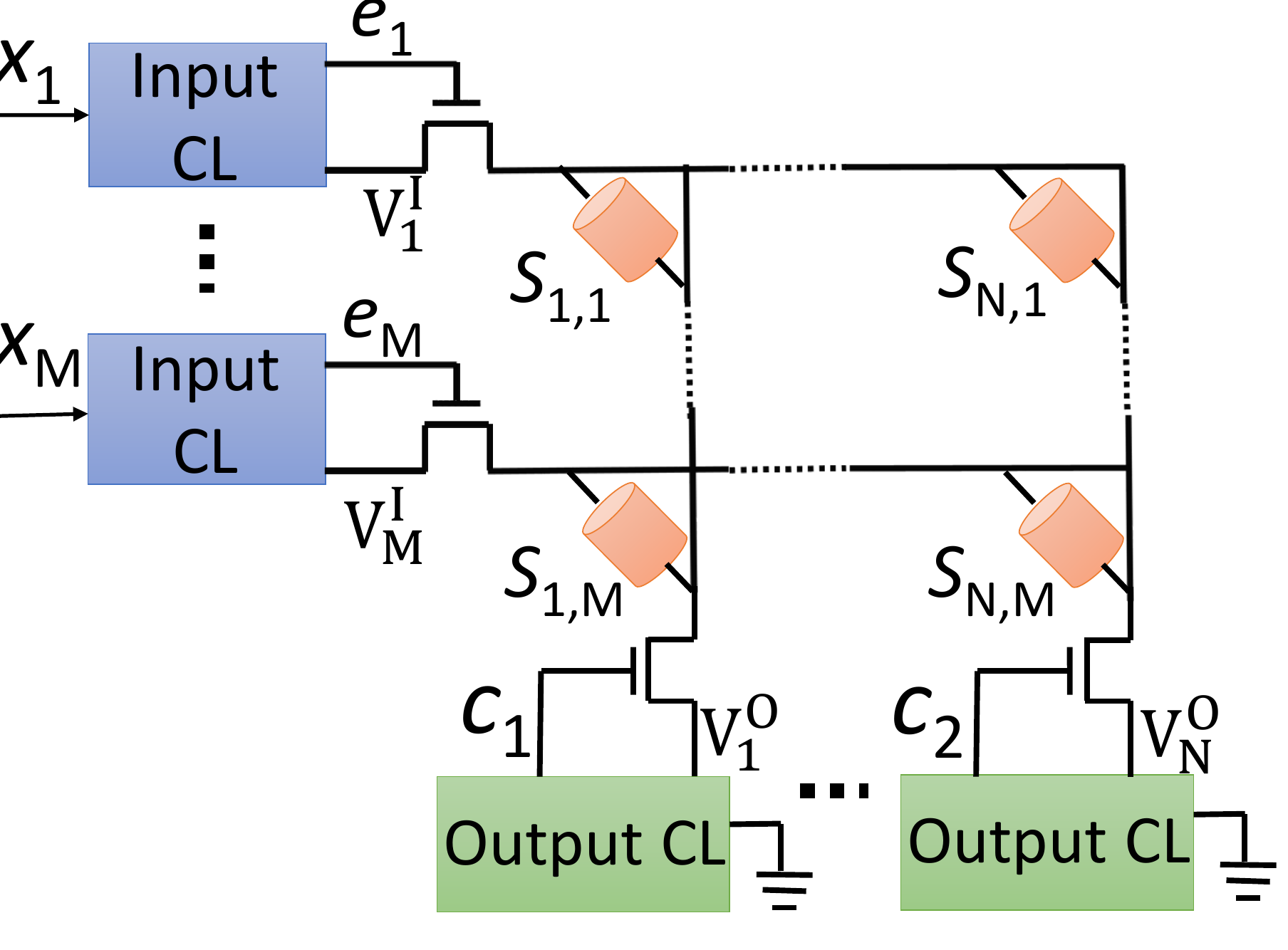}
    % Fully trimmed
    \label{crossbar_1r}
    }
    \par
\subfloat[]
{
\includegraphics[scale=0.29,trim = 5cm 11.2cm 14cm 1.5cm , clip]{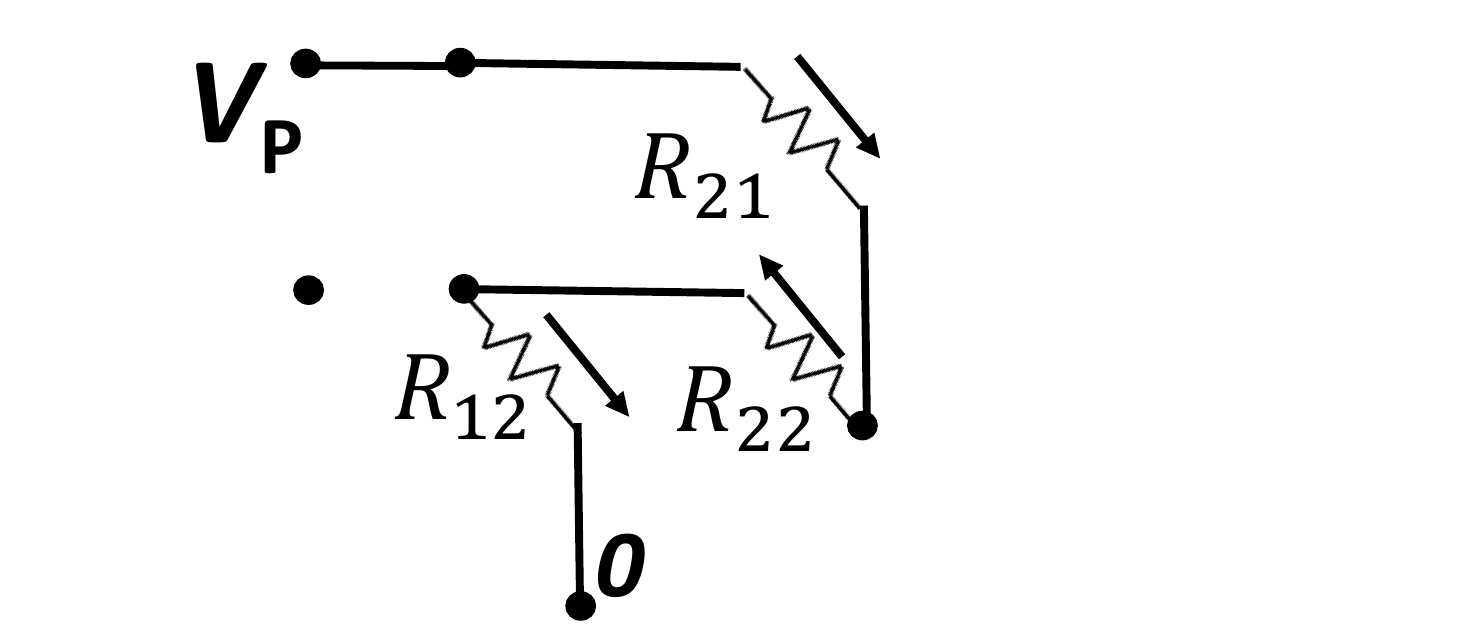}
\label{sneak_4phase_equiv}
}
\end{minipage}
\caption{\small (a) and (b) Alternate current paths in the 1R structure with 2-phase write strategy - (a) A $2\times 2$ crossbar. (b) Its equivalent circuit in write phase 1 with $c_1 = V_{DD}, c_2=0, V^O_1 = 0, V^I_1 = V_P, V^I_2 = V_{AP}$. (MTJ synapses shown as resistors). (c) Schematic of the proposed 1R Architecture for MTJ crossbar, (d) The equivalent circuit in phase 1 with 4-phase writing.} 
%(c) A passive column with $M$ inputs, $r$ of them being positive. The worst-case situation leading to an erroneous (d) $P\rightarrow AP$ switch, and (e) $AP\rightarrow P$ switch}
\label{sneak_1R_arch}
\end{figure}

Let's analyze the impact of sneak paths on the 1R crossbar with the 2-phase update strategy used previously. We first demonstrate the presence of sneak paths with a small example. Fig \ref{sneak_1R_arch}\subref{sneak_2phase_demo} shows a $2\times2$ crossbar with transistors only at the output terminals (to choose columns to be written in any particular phase). Assume without loss of generality that a certain input $x$ with $x_1>0, x_2<0$ produced errors $\delta_1>0,\delta_2<0$ at the outputs. The equivalent circuit during write phase 1 is drawn in fig. \ref{sneak_1R_arch}\subref{sneak_2phase_equiv}. It depicts the currents through the synapses, with the ones through $S_{21}$ and $S_{22}$ being undesired. These \textit{may} falsely switch $S_{21}$ from $P\rightarrow AP$ and $S_{22}$ from $AP\rightarrow P$ if they are in $P$ and $AP$ states respectively.

We now state a worst-case scenario for a crossbar with $M$ inputs. If $M$ is large, analysis using Kirchhoff's Current Law shows that the potential difference across an MTJ synapse could go as high as $(V_P - V_{AP})$. The current through such an MTJ, if in the $P$ state, is $I = (V_P - V_{AP})/R_P$ and is high enough (recall $V_{AP} < 0$) to switch it from $P\rightarrow AP$. In the other extreme case, a potential difference of $(V_{AP} - V_{P})$ leading to current $I = (V_{AP} - V_{P})/R_{AP}$ through an MTJ in the $AP$ state will switch it from $AP\rightarrow P$.

%The (worst-case) situation is exactly the same in phase 2. 
It is also necessary to mention an average (expected) case. Here these currents reduce to $I = (V_P - V_{AP})/2R_P$ and  $I = (V_{AP} - V_{P})/2R_{AP}$, respectively, which are half of those found previously, but still have some probability of switching MTJs (because these currents are roughly the same as $V_P/R_P$ and $V_{AP}/R_{AP}$). Thus, chances of unwanted flips of MTJs are quite significant, which calls for some modification in the circuit and/or in the programming method. 
%where each $x_i > 0 $ with probability 0.5 and a synapse is equally likely to be in the $P$ and $AP$ states. In such a scenario, 

\begin{table}[b]
    \setlength{\tabcolsep}{0.5mm}
    \renewcommand{\arraystretch}{1}
    \captionsetup{font={stretch=0.7}}
    \centering
    \begin{tabular}{|c|c|c|c|c|c|c|}
        \hline
         & Input & Error & $e_i$ & $V^I_i$ & $c_j$ & Switch \\
        \hline
        Phase 1 & $x>0$ & $\delta >0$ & $u(x_i)V_{DD}$ & $u(x_i)V_P$ & $u(\delta_j)V_{DD}$ & $P\rightarrow AP $ \\
        \hline
        Phase 2 & $x<0$ & $\delta >0$ & $u(-x_i)V_{DD}$ & $u(-x_i)V_{AP}$ & $u(\delta_j)V_{DD}$ & $AP\rightarrow P $ \\
        \hline
        Phase 3 & $x>0$ & $\delta <0$ & $u(x_i)V_{DD}$ & $u(x_i)V_{AP}$ & $u(-\delta_j)V_{DD}$ & $AP\rightarrow P $ \\
        \hline
        Phase 4 & $x<0$ & $\delta <0$ & $u(-x_i)V_{DD}$ & $u(-x_i)V_P$ & $u(-\delta_j)V_{DD}$ & $P\rightarrow AP $ \\
        \hline
    \end{tabular}
    \caption{4-phase weight update for the 1R configuration 
    in fig \ref{sneak_1R_arch}(c): Condition on input and error for a synapse to be updated, along with the control signals ($e$, $c$) and write voltages $(V^I)$, for each phase}
    \label{update_4phase}
\end{table}

\vspace{-1.5mm}
\subsubsection{\textbf{Four-phase Update:}}
\label{four_phase_update}
The large sneak currents in the 2-phase writing strategy, potentially resulting in false switching, is due to the high potential difference $V_P - V_{AP}$ between input terminals having different signs of inputs. One simple way to mitigate this issue is to further split the 2 phases of weight update so that, in a given phase, only rows having the same sign of input are updated at a time. This is equivalent to first clustering the columns according to the sign of $\delta$, and then further clustering the rows according to the sign of $x$. This proposed 4-phase writing scheme would require additional transistors to choose the rows to be updated in a given phase as shown in fig \ref{sneak_1R_arch}\subref{crossbar_1r}. It is summarized in Table \ref{update_4phase} where each phase will have the same duration $T_{wr}$; thus the total time for updating the crossbar is doubled to $4T_{wr}$. Note that this is still $O(1)$ time.

Let us now see how bad the issue of sneak-path leakage is with this strategy. Fig \ref{sneak_1R_arch}\subref{sneak_4phase_equiv} shows the equivalent circuit for the $2\times 2$ crossbar with the same set of assumptions (only synapses providing alternate current paths are shown). For an $M\times N$ crossbar, in the worst-case scenario, sneak currents could be $V_P/R_P$ and $V_{AP}/R_{AP}$, and can still result in false switching. This follows intuition as the potential difference between an input terminal and an output terminal is at most $V_P$ or $V_{AP}$. However, in the average case, the sneak current values are found to be only $V_P/3R_P$ and $V_{AP}/3R_{AP}$. These currents are small, and do not have the potential to cause undesired switching as is evident from the parameters listed in table \ref{params} and the range of values of $V_P$ and $V_{AP}$. Hence, the 4-phase writing scheme significantly reduces the incidences of undesired switching at a small cost of increase in the duration of the write phase. As we shall see, this trade-off is not only worth but also necessary for satisfactory performance of the training process.

\vspace{-2mm}
\subsection{Multi-Layer NNs} \vspace{-0.5mm}
\label{MLNN}
Multi-layer NNs can be implemented on cascaded crossbars (each representing one layer) with the output of one fed as the input to the next. It is pretty straightforward to implement the backpropagation algorithm on such a structure. Consider a 2-layer NN with weight matrices $W_1$ and $W_2$. For an input $x$, the final output $y_2$ is given as
\begin{equation}
\begin{aligned}
    y_2 = f(a_2) = f(W_2y_1) \ \ where \ \ 
    y_1 = f(a_1) = f(W_1x)
\end{aligned}
\end{equation}

If $\delta_2$ is the error of the second layer (output), then that of the first layer (hidden) is $\delta_1 = ({W_2}^T\delta_2) \times{f}^{\prime}(a_1)$ where ${f}^{\prime}$ is the derivative of activation function $f$, and $\times$ represents a component-wise product. This operation can be done on the crossbar (of the output layer) itself by reversing the roles of its input and output terminals: $\delta_2$ is now fed as the input and out comes ${W_2}^T\delta_2$, which, when multiplied by ${f}^{\prime}(a_1)$, gives $\delta_1$ as the error to be used for updating the weights of the hidden layer.

For the MTJ crossbar NN we described, during forward propagation, the total duration of the read phase would be $nT_{rd}$ for an $n$-layer NN. Backpropagation of errors to hidden layers would require an extra $T_{rd}$-long read phase for each such layer, during which the error at (the output of) a layer is fed as an input to its crossbar to obtain the error at its preceding layer. Lastly, all the layers can be updated simultaneously (in $2T_{wr}$ or $4T_{wr}$ time, as per the architecture).

Further, it must be mentioned that a large layer in an NN could be split into multiple crossbars, some of  which which share inputs or outputs. All these crossbars can still be read and written in parallel, thanks to the locality of the weight update operations.

%Variability in the properties of the MTJ from device-to-device may result in unexpected behavior. \textit{Recall that it is due to this reason that even the highly accessible 1T1R crossbar needs in-situ training.}

\vspace{-2mm}
\section{Experimental Setup and Results}

To see how successfully the MTJ crossbar NNs can be trained in-situ, we performed system level simulations by modeling the functionality of the crossbar architecture in MATLAB and training it on some datasets with supervised learning. To capture the MTJ device parameters, we used an HSPICE model \cite{kim2015technology} and included thermal fields in its LLG equations for obtaining the stochastic switching characteristics \cite{sengupta2016probabilistic}. Certain device parameters used in and obtained from this model were then incorporated into the simulations of the crossbar.

The performance of the neural network was evaluated in the following scenarios (code-named for further reference). All training processes used the Mean Square Error cost function and neurons had the \textit{tanh} activation function.
\begin{enumerate}[leftmargin=*,topsep=1pt, partopsep=0pt, parsep=0pt, itemsep=0pt,]
    \item \textbf{RV:} We first train and evaluate a neural network with \uline{r}eal-\uline{v}alued weights in MATLAB. Binary quantization step ($b$) is obtained from this trained network as shown in sec. \ref{binarization}.
    \item \textbf{DP:} Suitable binary weights are obtained by doing probabilistic learning in software on a binary network. Then a 1T1R crossbar and a 1R crossbar are \uline{d}eterministically \uline{p}rogrammed to these weights. We see the effect of device variations on the former, and of alternate current paths and resulting false switchings on the latter.
    \item \textbf{ST:} An MTJ synaptic crossbar is modeled and \uline{s}tochastically \uline{t}rained in-situ using the linear model of stochastic weight update described in sec. \ref{stochastic_update} for the
    \begin{enumerate}[topsep=1pt, partopsep=0pt, parsep=0pt, itemsep=1pt]
        \item 1T1R architecture, with the 2-phase write strategy (sec. \ref{1t1r_arch}).
        \item 1R architecture, with both the 2-phase (to see the effects of sneak currents) and the 4-phase update strategies (sec. \ref{1R_arch}). 
        %With the former, node voltages of output terminals not connected to the output CLs (that is, columns not being updated) were calculated using eqn. (\ref{node_voltages}). Whereas for the latter, a mesh analysis of the crossbar was required and node voltages at both (unconnected) input and output terminals were obtained by solving a system of linear equations.
    \end{enumerate}
    \item \textbf{DV:} \uline{D}evice \uline{v}ariations of different extent are introduced in the stochastic training of both the 1T1R and 1R crossbars. It  reflects in the variations in the resistance of the $P$ and $AP$ states, which usually doesn't exceed $10\%$ as per experiments \cite{worledge2010switching}. 
\end{enumerate}

%We first train and evaluate a neural network with real-valued weights in MATLAB with Mean Square Error cost function. Binary quantization step ($b$) is obtained from this trained network as shown in sec. \ref{binarization}. Also, a binary network with weights $W_b = b\hat{W}$ is evaluated to show the loss in network accuracy due to binarization. Next, an MTJ synaptic crossbar with 1T1R architecture is modeled in MATLAB and trained using the linear model of stochastic weight update (sec. \ref{stochastic_update}) with the 2-phase write strategy described in sec. \ref{Crossbar_arch}. The 1R architecture is similarly modeled and trained with both the 2-phase update strategy (to see the effects of sneak currents) and the 4-phase. With the former, node voltages of output terminals not connected to the output CLs (that is, columns not being updated) were calculated using eqn. \ref{node_voltages}. For the latter, a mesh analysis of the crossbar was required and node voltages at both (unconnected) input and output terminals were obtained by solving a system of linear equations.
We use the following datasets for evaluation.

\textbf{SONAR, Rocks vs Mines}\cite{Lichman:2013}: Three different NN architectures are considered - one with 1 layer (1L), and two with 2 layers having 15 and 25 hidden neurons respectively, and named 2L15 and 2L25. They were trained, and then tested on 104 samples of the test dataset.

\textbf{MNIST Digit Recognition}\cite{lecun1998gradient}: Three 2-layer networks of 50, 100 and 150 hidden units respectively and a 3-layer network of 50+25 hidden units were evaluated on the 10000 images of the test dataset.

\textbf{Wisconsin Breast Cancer (Diagnostic)}(WBCD)\cite{Lichman:2013}:  A single-layer network (1L) and 2 two-layer networks (2L10 and 2L20) were considered, and the test dataset had 200 samples.

\begin{figure}[t]
    \centering
    \captionsetup[subfloat]{margin=0pt}
    \subfloat[On SONAR for 2L15 network]
    {
    \includegraphics[scale=0.37,trim = 5cm 10.1cm 5.5cm 10.7cm , clip]{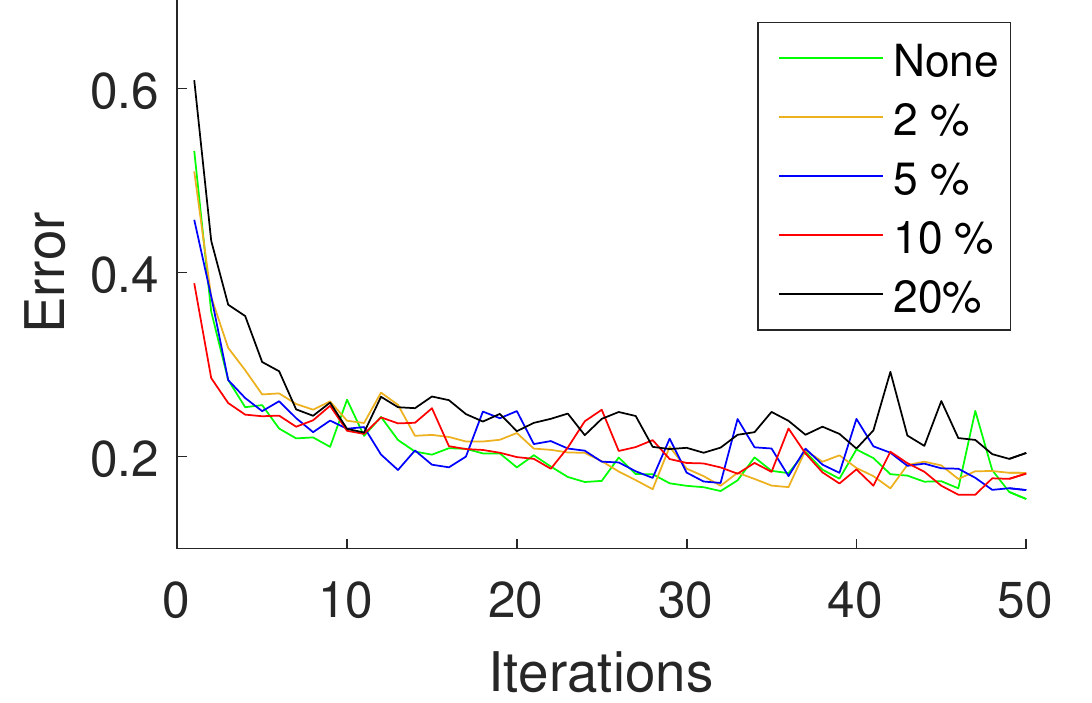}
    %Fully trimmed
    \label{sonar_variation}
    }
    %\hfill
    \captionsetup[subfloat]{margin=10pt}
    \subfloat[On MNIST for the 3L network]
    {
    \includegraphics[scale=0.41,trim = 5.5cm 10.5cm 6.0cm 11.0cm , clip]{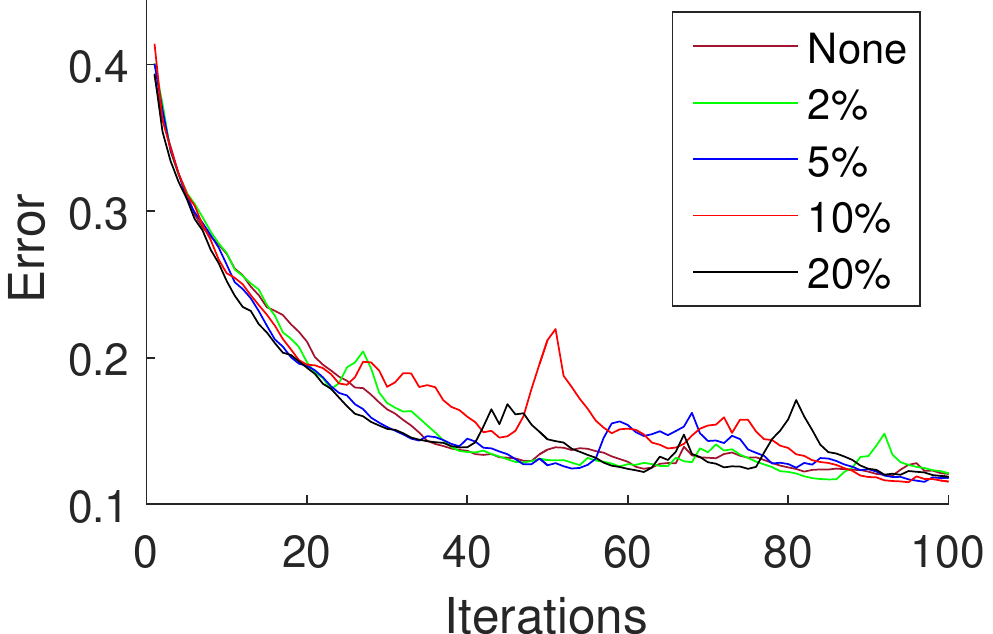}
    % Fully trimmed
    \label{mnist_variation}
    }
    \caption{Training error with different extents of device variations on the 1R crossbar architecture for 2 datasets.}
    \label{variation_figures}
\end{figure}

\begin{figure}
    \centering
    \captionsetup[subfloat]{margin=0pt}
    \subfloat[On SONAR for 2L15 network]
    {
    \includegraphics[scale=0.36,trim = 5cm 10cm 5.5cm 10cm , clip]{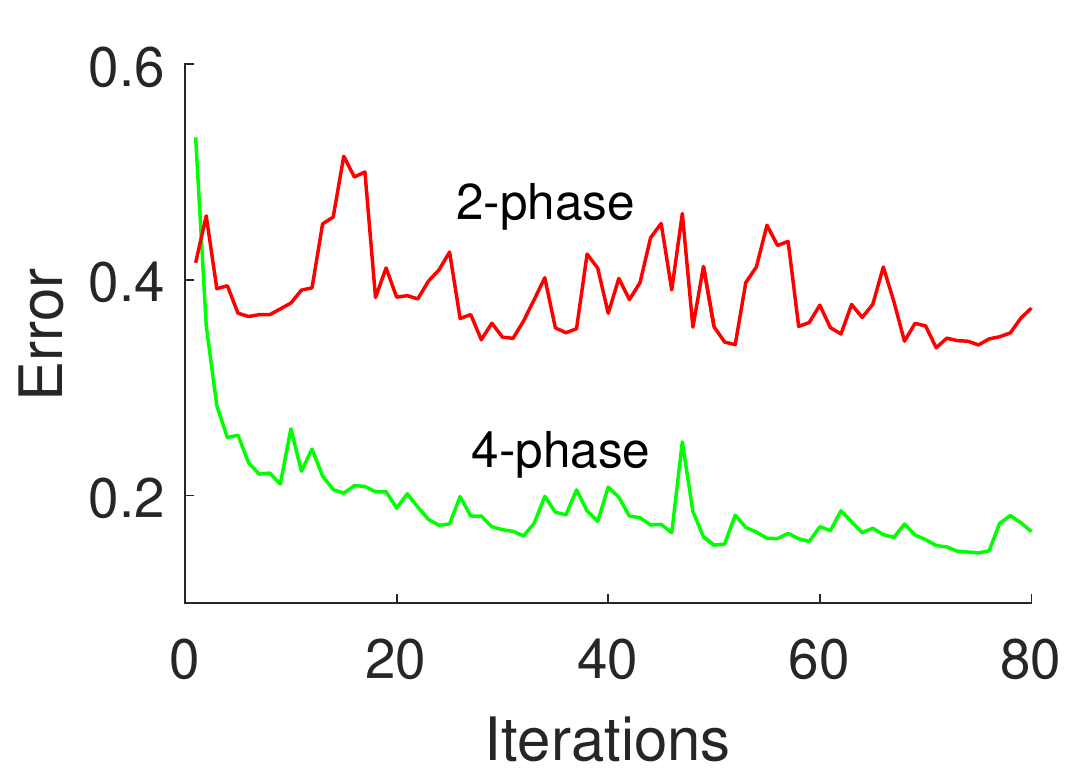}
    % Fully trimmed
    \label{sonar_compare}
    }
    %\hfill
    \captionsetup[subfloat]{margin=10pt}
    \subfloat[On MNIST for 2L100 network]
    {
    \includegraphics[scale=0.46,trim = 6.0cm 10.8cm 6.5cm 11.0cm , clip]{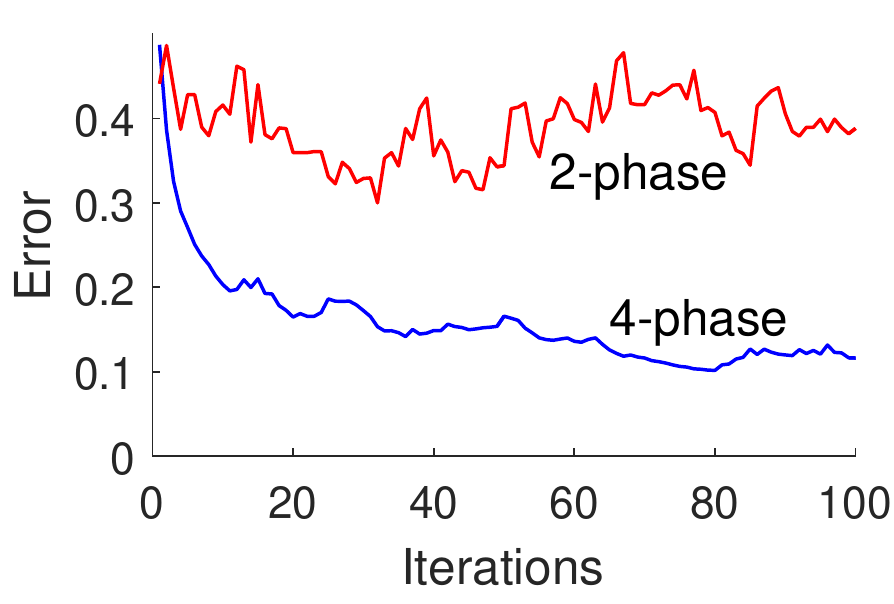}
    % Fully trimmed
    \label{mnist_compare}
    }
    \caption{Comparison of error during training of the 1R crossbar with 2-phase and 4-phase update schemes for 2 datasets. No variations assumed.}
    \label{compare_phase}
\end{figure}

Table \ref{accuracy_ideal} summarizes the accuracy obtained with these networks under the different training scenarios mentioned above. The effect of device variations of different extents on the in-situ stochastic training is highlighted for some of the networks in table \ref{accuracy_variation}, with fig. \ref{variation_figures} plotting the mean square error as the training progresses for the 1R crossbar. Additionally, fig. \ref{compare_phase} compares the error for the two write strategies. It doesn't converge with the 2-phase writing scheme due to higher instances of undesired weight changes, but does so with 4 phases. 

\begin{table}[t]
\setlength{\tabcolsep}{0.7mm}% Horizontal space between cell walls and cell text
    \renewcommand{\arraystretch}{0.9} % Vertical width of a cell
    \captionsetup{font={stretch=0.5}}
    %\centering
    \small
    {
    \begin{tabular}{|cc||c|c|c||c|c|c|c||c|c|c|}
        \hline
        \multicolumn{2}{|c|}{Dataset}  & \multicolumn{3}{c||}{SONAR} & \multicolumn{4}{c||}{MNIST} & \multicolumn{3}{c|}{WBCD} \\
         \hline
         \multicolumn{2}{|c|}{Network} & 1L & 2L15 & 2L25 & 2L50 & 2L100 & 2L150 & 3L & 1L & 2L10 & 2L20 \\
         \hline 
         \multicolumn{2}{|c|}{RV} & 16.4 & 12.8 & 11.9 & 9.87 & 7.34 & 6.44 & 7.25 & 8.35 & 7.40 & 7.10 \\
         \hline
         \multirow{2}{*}{DP} & \multicolumn{1}{|c|}{1T1R} & 19.2 & 15.2 & 14.3 & 13.50 & 10.89 & 9.55 & 10.45 & 9.85 & 8.30 & 8.55 \\
         \cline{2-12}
         & \multicolumn{1}{|c|}{1R} & 46.8 & 41.4 & 42.7 & 39.42 & 36.10 & 37.92 & 40.48 & 24.95 & 27.60 & 23.65 \\
         \hline
         \multirow{2}{*}{ST} & \multicolumn{1}{|c|}{1T1R} & 18.4 & 14.2 & 13.6 & 12.69 & 10.18 & 8.96 & 9.71 & 9.20 & 7.70 & 8.05 \\
         \cline{2-12}
          & \multicolumn{1}{|c|}{1R} & 18.3 & 14.5 & 14.0 & 12.72 & 10.20 & 9.03 & 9.66 & 9.40 & 7.85 & 7.95 \\
          \hline
    \end{tabular}
    }
    \caption{Classification error rates for the 3 datasets (on the test samples) with various NN and crossbar architectures under different training scenarios. Here, ST-1R crossbar used 4-phase update. Ideal devices assumed for all except DP-1T1R, where $10\%$ variation was considered. SONAR and WBCD figures are average of 10 runs. MNIST and WBCD figures are in \%}
    \label{accuracy_ideal}
\end{table}

\begin{table}[t]
\setlength{\tabcolsep}{0.7mm}% Horizontal space between cell walls and cell text
    \renewcommand{\arraystretch}{0.9} % Vertical width of a cell
    \captionsetup{font={stretch=0.5}}
    %\centering
    \small
    {
    \begin{tabular}{|c|c|c|c|c||c|c|c|c||c|c|}
        \hline
        \multicolumn{1}{|c|}{Dataset}  & \multicolumn{4}{c||}{SONAR} & \multicolumn{4}{c||}{MNIST} & \multicolumn{2}{c|}{WBCD} \\
         \hline
        \multicolumn{1}{|c|}{Network} & \multicolumn{2}{c|}{1L} & \multicolumn{2}{c||}{2L15} & \multicolumn{2}{c|}{2L100} & \multicolumn{2}{c||}{3L} & \multicolumn{2}{c|}{2L20} \\
        \hline
        Variation & 1T1R & 1R & 1T1R & 1R & 1T1R & 1R & 1T1R & 1R & 1T1R & 1R \\
        \hline
        2\% & 18.5 & 18.4 & 14.4 & 14.7 & 10.27 & 10.22 & 9.67 & 9.73 & 8.10 & 8.05\\
        \hline
        5\% & 18.7 & 18.7 & 14.7 & 14.8 & 10.28 & 10.29 & 9.78 & 9.80 & 8.25 & 8.30\\
        \hline
        10\% & 19.0 & 19.1 & 15.1 & 15.1 & 10.33 & 10.43 & 9.86 & 9.91 & 8.30 & 8.40\\
        \hline
        20\% & 19.3 & 19.5 & 16.0 & 15.9 & 10.42 & 10.72 & 10.15 & 10.28 & 8.60 & 8.75 \\
        \hline
        %DP(1T1R) & 19.2 & -- & 15.2 & -- & 10.89 & -- & 10.45 & -- & 8.55 & -- \\
        %\hline
    \end{tabular}
    }
    \caption{Misclassification rates with stochastic training (ST) of 1T1R and 1R architectures under different levels of device variations (DV).}
    \label{accuracy_variation}
\end{table}

It is evident from these results that 
\begin{itemize}[itemindent=7pt,leftmargin=0pt,topsep=1pt, partopsep=0pt, parsep=0pt, itemsep=0pt]
    %\item An MTJ synaptic crossbar without access transistors suffers from severe loss in performance when it is programmed deterministically due to undesired weight changes arising from alternate current paths.
    \item When an MTJ synaptic crossbar without access transistors is  stochastically trained in-situ (ST-1R), it shows classification accuracy only slightly lower (about $3\%$ at worst) than when the same network is trained in software with real-valued weights (RV, which can be considered to be the best achievable). However, it brings about significant improvement (up to $30\%$) in accuracy over a deterministically programmed crossbar (DP-1R) since the latter suffers from undesired weight changes arising from alternate current paths.
    \item In-situ training also benefits the crossbar with transistors (ST-1T1R against DP-1T1R) in the presence of device variations by slightly improving accuracy (by about $0.5\%-1\%$).
    \item It is possible to compensate for the loss in accuracy due to use of a binary network by increasing the size of the network (adding more hidden layers and/or neurons).
    \item Further, the trained crossbar has robustness even in the face of device variations, owing primarily to the fault-tolerant nature of NN and its learning algorithms. As can be seen in table \ref{accuracy_variation}, increase in misclassification rates remain within $2\%$ even with $20\%$ variation.
\end{itemize}

The accuracy degradation of $2\%-3\%$ that we achieve (on going from RV to ST) is comparable to the $3.73\%$ reported by \cite{zhang2016stochastic} and the $0.8\%-3.5\%$ in \cite{vincent2015spin}. However, it must be mentioned and emphasized that any comparison is fair only if they are on the same dataset and network architecture. The benefit of using in-situ training can also be seen when we compare our work with that of \cite{zhang2016all} (which performs offline learning). On the MNIST 2L100 network, we obtained an error rate of $10.20\%$, whereas \cite{zhang2016all} had a much higher value of $30\%$ on the same network, although it must be mentioned that the latter were at a disadvantage due to linear activation units.

 %Performance under different scenarios and variations for different networks architectures are highlighted in table \ref{mnist}. Fig. \ref{mnist_figures} \subref{mnist_variation} and \subref{mnist_arch} show the convergence of error during training for different extents of device variations and crossbars, respectively. We conclude from the latter that, although the error on a 1R crossbar fluctuates (due to leakage currents), it converges to nearly the same levels as that of the 1T1R version.

\vspace{-2mm}
\section{Conclusion}

In this work, we show how MTJ crossbars representing weights of an ANN can be trained in-situ by exploiting the stochastic switching properties of MTJs and performing weight updates in a way akin to gradient descent. We demonstrate how the learning algorithm can be implemented on crossbars with and without transistors. Results show these stochastically trained binary networks can achieve classification accuracy almost as good as that of those trained in software and implemented on processors. This paves the way for the attainment of highly scalable neural systems in the future capable of performing complex applications.

\vspace{-2.0mm}
\bibliographystyle{ACM-Reference-Format}
\bibliography{main} 

\end{document}